%% file: main.tex
\documentclass[preprint]{elsarticle}
\usepackage[margin=1in]{geometry}

\usepackage[bookmarksnumbered]{hyperref} 
\usepackage[mode=buildnew]{standalone}

\usepackage{amssymb,amsmath,amsthm,amsbsy, mathtools}
\usepackage{systeme}
\usepackage{physics}
\usepackage{graphicx}
\usepackage{epstopdf}
\usepackage{epsfig}
\usepackage{multicol}
\usepackage{multirow}
\usepackage{fontawesome5}
\usepackage{algorithm,algorithmic}
\usepackage{bm}
\usepackage[capitalise]{cleveref}
\usepackage{enumerate}
\usepackage{setspace}
\usepackage{color}
\usepackage{resizegather}
\usepackage{booktabs}
\usepackage{tabularx}
\usepackage{placeins}
\usepackage{numprint}
\npdecimalsign{.}
\nprounddigits{3}

\usepackage{tikz}
\usepackage{subcaption}
\usepackage[mathlines,running]{lineno}
\usepackage[normalem]{ulem}

\input{variables}
\input{Figures/figure_config}

\begin{document}
\makeatletter
\def\ps@pprintTitle{%
  \let\@oddhead\@empty
  \let\@evenhead\@empty
  \let\@oddfoot\@empty
  \let\@evenfoot\@oddfoot
}
\makeatother


\begin{frontmatter}

\title{VENI, VINDy, VICI: a generative reduced-order modeling framework with uncertainty quantification}

\author[mox]{Paolo Conti\corref{contrib}}
\ead{paolo.conti@polimi.it}
\author[simtech]{Jonas Kneifl\corref{contrib}} 
\ead{jonas.kneifl@itm.uni-stuttgart.de}
\author[mox]{Andrea Manzoni}
\ead{andrea1.manzoni@polimi.it}
\author[dica]{Attilio Frangi}
\ead{attilio.frangi@polimi.it}
\author[simtech]{J\"org Fehr}
\ead{joerg.fehr@itm.uni-stuttgart.de}
\author[UW1]{Steven L. Brunton}
\ead{sbrunton@uw.edu}
\author[UW2]{J.~Nathan Kutz}
\ead{kutz@uw.edu}

\address[mox]{MOX -- Department of Mathematics, Politecnico di Milano, Milan, Italy}
\address[simtech]{Institute of Engineering and Computational Mechanics, University of Stuttgart, Stuttgart, Germany}
\address[dica]{Department of Civil Engineering, Politecnico di Milano, Milan, Italy}
\address[UW1]{Department of Mechanical Engineering, University of Washington, Seattle, USA}
\address[UW2]{Department of Applied Mathematics and Electrical and Computer Engineering, University of Washington, Seattle, USA}
\cortext[contrib]{These authors contributed equally to this work and are listed in alphabetical order.}

\begin{abstract}
Generative models are transforming science and engineering by enabling efficient synthetization and exploration of new scenarios for complex physical phenomena with minimal cost. 
Although they provide uncertainty-aware predictions to support decision making, they typically lack physical consistency, which is the backbone of computational science.
Hence, we propose VENI, VINDy, VICI -- a novel \textit{physical generative framework} that integrates data-driven system identification into a probabilistic modeling approach to construct physically consistent and efficient eeduced-order models with uncertainty quantification. \\
First, VENI (Variational Encoding of Noisy Inputs) employs variational autoencoders to identify reduced coordinates from high-dimensional, noisy measurements. 
Simultaneously, VINDy (Variational Identification of Nonlinear Dynamics) extends sparse system identification methods by embedding probabilistic modeling into the discovery process.
Last, VICI (Variational Inference with Credibility Intervals) enables efficient generation of full-time solutions and provides uncertainty quantification for unseen parameters and initial conditions.
We demonstrate the performance of the framework across chaotic and high-dimensional nonlinear systems.
\end{abstract}


\end{frontmatter}

\section{Introduction}
Scientific computing offers a well-estabilished suite of simulation tools that span a wide range of applications in engineering and science, enabling the exploration of complex phenomena that would otherwise be intractable. 
These phenomena are typically governed by partial differential equations (PDEs), which provide an accurate mathematical framework for their description. 
However, solving PDEs with high accuracy can be computationally demanding, particularly for high-dimensional systems, complex geometries, or applications requiring repeated evaluations, such as uncertainty quantification (UQ) \cite{bui2008parametric,smith2024uncertainty} or PDE-constrained optimization \cite{manzoni2021optimal}. 

Reduced-order models (ROMs) have emerged as a promising solution to mitigate these challenges. They represent high-dimensional solutions within a lower-dimensional space, enabling efficient computations, while still retaining the essential accuracy \cite{HRS, QMN, benner2015survey}. 
Traditional ROM techniques rely on detailed knowledge of the governing equations, necessitating domain-specific expertise. This becomes a significant barrier, especially in complex fields like turbulent flow and climate modeling, where the governing equations may be incomplete or unknown.
In response, data-driven methods have gained popularity because they can create ROMs directly from observed data, bypassing the need for detailed physical insight. Recent techniques for extracting governing equations from data include dynamic mode decomposition \cite{schmid2010dynamic, brunton2022data} and deep learning approaches based on autoencoders \cite{lusch2018deep, champion2019data, chen2022automated,fries2022lasdi, conti2023reduced, bakarji2023discovering}, which combine dimensionality reduction with the identification of latent dynamics. However, while these methods improve generalization and predictive capabilities, they often lack robust UQ mechanisms, making them less reliable in noisy or data-limited environments.

Recently, generative models have demonstrated exceptional capabilities in addressing uncertainties and capturing complex patterns in high-dimensional data, e.g. in fields such as computer vision and natural language processing \cite{Devlin2018, Brown2020}. 
In the context of ROMs, generative models offer transformative potential by enabling efficient synthesis of new PDE solutions, enhancing robustness to noise, and naturally incorporating UQ.
Yet, their application in scientific computing remains limited due to the challenge of ensuring physical consistency and reliability. 

\begin{figure}[t!]
    \centering
    \includegraphics[width=\textwidth]{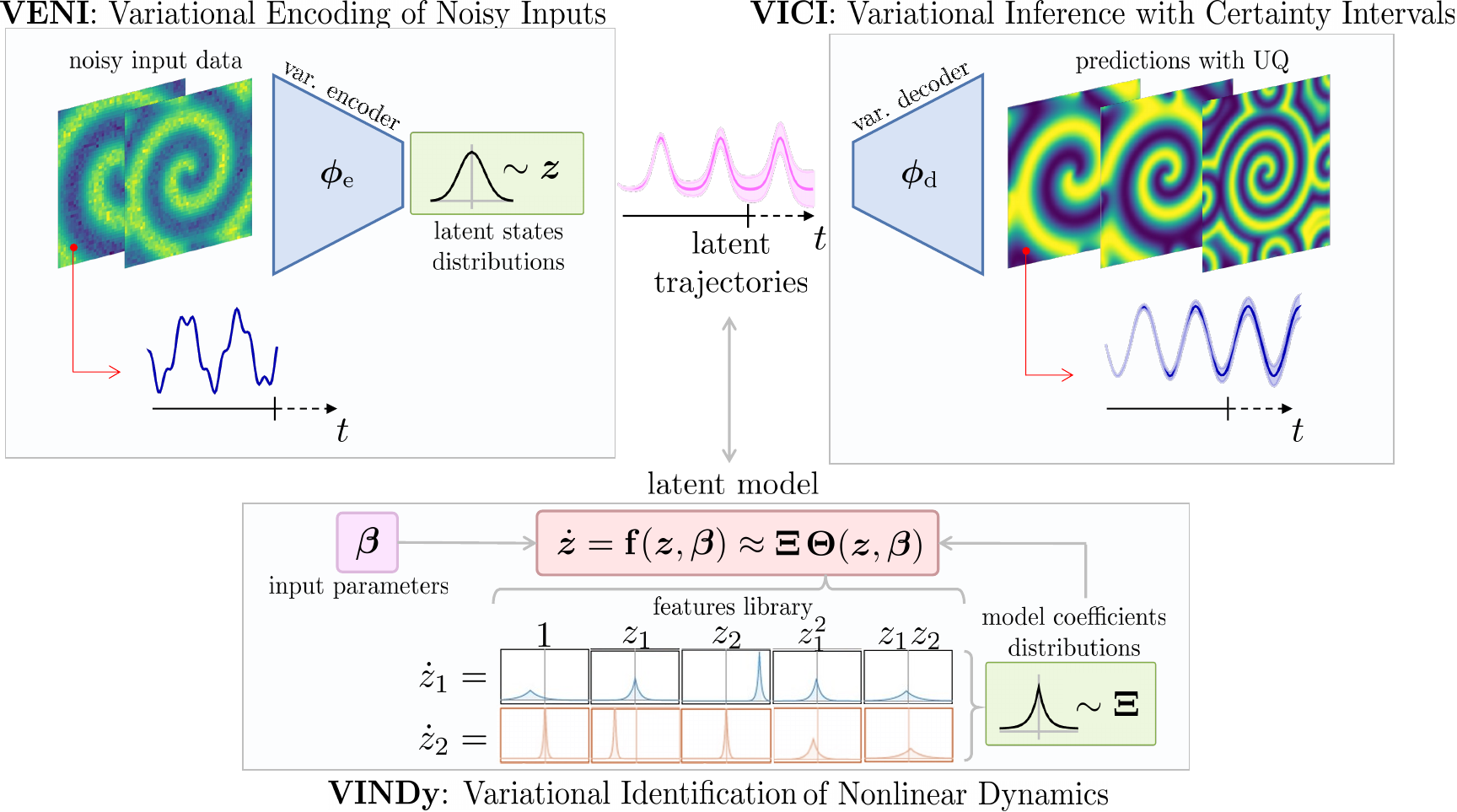}
    \caption
    {Overview of the VENI, VINDy, VICI procedure. High-dimensional, noisy data are mapped through a variational encoder to low-dimensional, latent random variables (VENI). Simultaneously, in a joint offline-training, the dynamics of the latent variables are learned by VINDy. Once this offline phase is concluded, noise-free, full-field solutions are computed through the online generative process together with the predictions uncertainty bounds (VICI).}
    \label{fig: overview}
\end{figure}

To overcome the lack of physical consistency in generative models, we propose a novel \textit{physical generative model} that seamlessly integrates dimensionality reduction, dynamical system identification, and UQ within a unified variational framework. 
%
Our approach embeds latent dynamics discovery directly within the generative model, promoting interpretability and physically meaningful representations of the dynamics using parsimony in sparse regression as a regularizer \cite{kutz2022parsimony}, while retaining the computational efficiency and scalability of variational methods.
By combining data-driven identification of reduced variables and governing equations with UQ, this hybrid framework positions our method as a critical advancement for ROMs, enabling automated, interpretable, and uncertainty-aware models that can reliably capture complex system dynamics.
%
%
The method consists of three steps: 
\begin{itemize}
    \item[\textit{(i)}] \textbf{VENI} (Variational Encoding of Noisy Inputs).
    A variational autoencoder (VAE) \cite{kingma2013auto} is employed to identify the distribution of reduced latent states  from high-dimensional, noisy snapshots. 
    VAEs promote continuity and smoothness in the latent space and, when combined with explicit inductive biases, can encourage more independent, disentangled reduced coordinates \cite{higgins2018towards, alemi2016deep}, while not guaranteeing identifiability in the fully unsupervised case \cite{Locatello2018}. VAEs demonstrated successful applications in dynamical system \cite{solera2024beta, simpson2024vprom, botteghi2022deep}.
    \item[\textit{(ii)}] \textbf{VINDy} (Variational Identification of Nonlinear Dynamics).
   This novel component extends sparse identification of nonlinear dynamics (SINDy) \cite{brunton2016discovering} to a variational framework in order to learn a probabilistic dynamical model in the latent coordinates.
    Given a library of candidate features, VINDy identifies a sparse subset of active function terms and combines them linearly to approximate the system dynamics. In contrast to SINDy, the multiplicative coefficients are parameterized as probability distributions which capture uncertainty in the dynamics.
\end{itemize}
VENI and VINDy are trained simultaneously, ensuring that the reduced coordinates learned by VENI enable VINDy to identify a physically meaningful dynamical model.
This integrated approach leverages variational inference to achieve efficient and scalable training while avoiding the computational overhead associated with alternative Bayesian  \cite{Hirsh2022sparsifying, gao2022bayesian, niven2024dynamical} or weak-form \cite{messenger2021weak, Wang2019Variational} methods.
\begin{itemize}
    \item[\textit{(iii)}] \textbf{VICI} (Variational Inference with Credibility Intervals). 
    VICI represents the online phase, where the probabilistic ROM generates approximate, full-state solutions and quantifies prediction uncertainty for unseen parameters and/or initial conditions. By leveraging the learned probabilistic representations, VICI ensures reliable extrapolation and provides uncertainty intervals for robust decision-making.
\end{itemize}
A schematic representation of the overall framework -- named VENI, VINDy, VICI~-- is illustrated in Fig.~\ref{fig: overview}.
Our strategy offers an effective solution by converting the challenge of UQ into an integrated optimization problem solvable with gradient-based methods. This approach naturally encodes UQ into the training process by adding suitable regularization terms to the loss function.  This ensures that the optimization efficiency of neural networks is maintained, avoiding the need for computationally intensive techniques like Monte Carlo methods.

Recent work has explored alternative approaches to modeling stochastic dynamics, such as coupling variational autoencoders with hyper-networks \cite{jacobs2023hypersindy} or applying stochastic variational inference with Gaussian processes \cite{course2023state}.  Additionally, some studies are leading the way in automatically discovering latent state variables  \cite{chen2022automated} and governing equations \cite{gao2022bayesian} from video data, although they typically rely on non-probabilistic encoding methods.
The present work makes the following contributions:
\begin{enumerate}
    \item We develop a novel data-driven system identification approach (VINDy) which is based on variational inference and incorporates UQ.
    \item We integrate dimensionality reduction (VENI), system identification (VINDy), and UQ into a unified optimization framework that can be efficiently trained using a variational approach.
    \item We demonstrate the potential of our approach on the Rössler system, illustrating its performance in a low-dimensional setting with varying noise intensities and sources on a chaotic system, and on  high-dimensional PDE benchmarks,  including MEMS (Micro Electro-Mechanical System) resonators and an unsteady PDE in fluid dynamics, consisting of a parametrized reaction-diffusion problem. 
\end{enumerate}
This paper is structured as follows. In \cref{sec: method} we detail the VENI, VINDy, VICI method in its components. We present how to perform the offline training and how to use the method to efficiently generate full-time solution. 
Then, we present the potential of the newly introduced VINDy method alone on the R\"ossler system, thus showcasing the methodology on a didactic, low-dimensional system with different noise intensities and sources (both model and measurement uncertainties).
Next we validate the performance of the complete framework through a series of high-dimensional PDE benchmarks, including a straight beam MEMS resonator, which is excited at different forcing amplitudes and frequencies, and an unsteady PDE in fluid dynamics, consisting in a parametrised reaction-diffusion problem. 
Results for these numerical tests are showcased in \cref{sec: results}, finally we discuss results and draw some concluding remarks in \cref{sec: conclusions}. 
The source code of the proposed method is made available in the public repository \url{https://github.com/jkneifl/VENI-VINDy-VICI}.

\section{Method}
\label{sec: method}

\subsection{Problem statement}
Our framework is designed to construct a physically constrained generative model for the efficient generation of solutions to nonlinear, parameterized, dynamical systems governed by partial differential equations, given a limited set of noisy snapshots of the system states. 
Specifically, we aim to compute uncertainty-aware predictions for dynamical systems expressed in the general form
\begin{equation}
    \begin{cases} 
    \dot{\states}(t; \param) = \vb{F}(t,\states(t; \param);\param), & t \in (0,T),\\ 
    \states(0;\param) = \states_0,
    \end{cases}
\label{eq: dy_eq_full}
\end{equation}
where $\states \in \mathbb{R}^{\stateDim}$ is the state of the system, $\dot{\states}$ its derivative with respect to time $t$ and $\param = [\beta_1, \ldots, \beta_p]^\top\in \mathbb{R}^{p}$ the vector collecting $p$ (possibly time dependent) parameters and/or forcing terms. 
The function $\vb{F}$ defines the dynamics of the physical system, evolving from the initial state $\states_0$.
As the state dimension $\stateDim$ is typically extremely large, system \eqref{eq: dy_eq_full} is denoted as full-order model (FOM).

We split the objective into two tasks: \textit{(i)} construct a generative model that allows us to reproduce full, spatial states $\states$ of the system, and \textit{(ii)} identify the system's governing dynamics $\mathbf{F}$ in a probabilistic framework, enabling predictions of state evolution under uncertainty.
However, solving these tasks in the high-dimensional space can be extremely involved and computationally demanding, 
while often the solution manifold is of a significantly lower dimension \cite{Fefferman2016}. 
Consequently, 
our reduced-order modeling technique reduces the problem's dimensionality and expresses \eqref{eq: dy_eq_full} in a new set of reduced coordinates
more optimally suited to represent the dynamics of the problem and also accounting for uncertainty quantification. 
More specifically, task \textit{(i)-(ii)} are addressed respectively by VENI, which employs VAEs to provide the latent representation and a generative model for the system's states $\states$, and VINDy, which identifies the governing dynamics in the new set of latent coordinates.

Finally, in the online phase called VICI, the temporal evolution of the system's solution is computed for new parameter/forcing values $\param$ and initial conditions $\states_0$, using variational inference on both the latent variable distribution and the dynamic model to ensure robust predictions and reliable uncertainty quantification.

\subsection{Variational Encoding of Noisy Inputs (VENI)}
\label{sec: VENI}
To generate the states $\states$ of the system, VENI aims to approximate their unknown probability distribution $p(\states)$ over $\mathbb{R}^\stateDim$, given a limited set of noisy snapshot data of the system $\dataset\coloneqq\{\states^{(i)}\}_{i=1}^{n_s}$.
Given the high dimensionality of the states and the limited amount of available data, this task is particularly challenging. 
We address it by leveraging variational autoencoders (VAEs), introduced in \cite{kingma2013auto} and comprehensively explained in \cite{yu2020tutorial}. 
VAEs allow variational inference and UQ for high-dimensional datasets ~\cite{abdar21review, Gundersen2021}, by approximating data distributions in a low-dimensional, latent space.
Moreover, using inductive bias in VAEs can promote disentanglement of latent features \cite{Locatello2018} and allows to control the way latent distributions are modeled through the selection of appropriate priors~\cite{gao2022bayesian, Hirsh2022sparsifying}.

Since the solution states $\states$ of dynamical systems governed by PDEs typically lie on a low-dimensional manifold embedded in the full-order space, we assume that each state $\states$ can be generated from a set of low-dimensional, latent variables $\latentState = [z_1, \ldots, z_\latentStateDim]^\top \in \mathbb{R}^\latentStateDim$, where $\latentStateDim \ll \stateDim$, through the \textit{decoder}
\begin{equation}
    \decoder(\cdot\,;\weights_{\decoder}): \R^\latentStateDim\rightarrow\mathcal{P},\text{ such that } \latentState\mapsto\decoder(\latentState;\weights_{\decoder})=~ p(\states|\latentState;\weights_{\decoder}),
\label{eq: decoder}
\end{equation}
where ~$p(\states|\latentState;\weights_{\decoder})$ belongs to a family~$\mathcal{P}$ of probability distributions over $\R^\stateDim$, parametrized by weights~$\weights_{\decoder}$. 
VAEs, in addition, feature an \textit{encoder}
\begin{equation}
    \encoder(\cdot\,;\weights_{\encoder}): \R^\stateDim\rightarrow\mathcal{Q},\text{ such that } \states\mapsto\encoder(\states;\weights_{\encoder})=~ \varPosterior(\latentState|\states;\weights_{\encoder}),
    \label{eq: encoder}
\end{equation}
with weights~$\weights_{\encoder}$, to generate values of those latent variables $\latentState$ from a preselected family of distributions~$\mathcal{Q}$ that are likely to have produced the snapshot $\states$ under consideration of a predefined prior~$\prior$. 

The objective to maximize the probability~$p(\states)$ of the present data is in general not tractable, making direct optimization infeasible. To circumvent this issue in VAEs, the Evidence Lower Bound (ELBO) \cite{kingma2013auto, yu2020tutorial}
\begin{align}
    \log \marginal - \KL(\varPosterior(\latentState|\states)\parallel\posterior) &= \expectation_{\latentState\sim \varPosterior}[\log p(\states|\latentState)] - \KL(\varPosterior(\latentState|\states) \parallel \prior)\label{eq: objective1}\\
    &= \expectation_{\latentState\sim \varPosterior}[\log\decoder(\latentState)] - \KL(\encoder(\states) \parallel \prior) \label{eq: objective2}
\end{align}
provides a computationally tractable alternative.
Here, $\KL$ describes the Kullback-Leibler divergence
\begin{equation}  
    {\KL\left(q(\latentState|\states)\parallel\posterior\right) \coloneqq \expectation_{\latentState\sim \varPosterior}\left[\log q(\latentState|\states)- \log\posterior\right]},
    \label{eq: KL}
\end{equation}
which is a measure to compare the proximity of two distributions.
By optimizing the right-hand side of~\eqref{eq: objective1}, we effectively maximize a lower bound on the intractable optimization goal.
The term $\expectation_{\latentState\sim \varPosterior}[\log p(\states|\latentState)]$ in \eqref{eq: objective1} ($\expectation_{\latentState\sim \varPosterior}[\log\decoder(\latentState)]$ in \eqref{eq: objective2}, respectively) expresses the reconstruction loss. 
It ensures that the log-probability of~$\states$ given~$\latentState$ drawn from the approximated posterior $\varPosterior(\latentState \vert \states)$ is maximized, \i.e. that the data are reconstructed well by the decoder $\decoder$ from a sample in the reduced space. 
The second term ensures that the approximated posterior $\varPosterior(\latentState|\states)$ is pushed closer to the prior, \i.e. that reduced coordinates identified by the encoder $\encoder$ respect the assumed prior over the latent variables~$\prior$.

The classes of {encoder} and {decoder} (in particular, the family of distributions $\varPosteriorFamily$ and $\mathcal{P}$, respectively) must be chosen in such a way that their weights, $\weights_{\encoder}$ and $\weights_{\decoder}$ can be optimized using standard machine learning optimization techniques.
In this work, the choice falls on Gaussian distribution, resulting in a decoder
\begin{equation}
 \decoder(\latentState;\weights_{\decoder}) =p(\states|\latentState;\weights_{\decoder}) = \NormalDist\left(\vaeMean(\latentState;\weights_{\decoder}),\sigma^2 \bm{I}_{\stateDim}\right),
\label{eq: decoder_dist}
\end{equation}
where the mean $\vaeMean\in\R^\stateDim$ is implemented by multi-layer, feed-forward, neural network, while the covariance is an isotropic matrix with variance $\sigma^2$ (scalar hyperparameter). 
Even though any probability distribution continuous in $\weights_{\decoder}$ can be employed, the advantage of using an isotropic Gaussian distribution is that, the term $\expectation_{\latentState\sim \varPosterior}[\log \decoder(\latentState)]$ in \eqref{eq: objective2} is proportional to the squared distance between the neural network reconstruction $\hat{\states}=\vaeMean(\latentState;\weights_{\decoder})$ and the original data $\states$ \cite{yu2020tutorial}. In this way, the decoder can be trained on the classical reconstruction loss of standard autoencoders: $\norm{\states - \hat{\states}}_2^2$.
 
The candidate posterior distributions $\varPosteriorFamily$ of the encoder $\encoder$ are assumed to belong to the same family as the prior distribution of the latent variables. 
If we assume independent, standard Gaussians, \i.e. $\prior=~\NormalDist(\bm{0}, \bm{I}_\latentStateDim)$ as priors for the latent variables, the encoders' posterior is defined as
\begin{equation}
    \encoder(\states;\weights_{\encoder}) =q(\latentState|\states\mathcal;\weights_{\encoder})=\NormalDist\left(\hat{\latentState}(\states;\weights_{\encoder}), \text{diag}(\vaeVar_{\encoder}(\states;\weights_{\encoder}))\right).
\label{eq: encoder_dist}
\end{equation}
Here, both the mean~$\hat{\latentState}~\in~\Rdim^\latentStateDim$ and the variances~$\vaeVar_{\encoder}\in~\Rdim^{\latentStateDim}_+$ are implemented as output of a multi-layer, feed-forward neural network. The notation $\text{diag}(\vaeVar_{\encoder})\in\Rdim^{n\times n}$ indicates the diagonal matrix having $\vaeVar_{\encoder}$ on the main diagonal. This choice is advantageous, as the term $\KL(\encoder(\states)|\prior)$ in \eqref{eq: objective2} can be then written in closed-form for distributions such as Gaussian and Laplacian (see \ref{app: prior}).

\subsection{Variational Identification of Nonlinear Dynamics (VINDy)}
\label{sec: VINDy}
Despite the VAEs’ capabilities in identifying low-dimensional variables from which the high-dimensional
distribution can be generated, they do not account for time or dynamics. 
VINDy aims to describe how the system states evolve in time by identifying a dynamical system of the form
\begin{equation}
    \begin{cases} 
    \dot{\latentState}(t; \param) = {\vb{f}}(t,\latentState(t; \param);\param), & t \in (0,T),\\ 
    \latentState(0;\param) = \latentState_0,
    \end{cases}
\label{eq: dy_eq_reduced}
\end{equation}
in the latent coordinates~$\latentState\in\Rdim^\latentStateDim$ defined by the VAE, where $\latentState_0 \sim \encoder(\states_0)$ and $\dot{\latentState}$ represents the latent states'~time derivatives.
In particular, we are interested in identifying the unknown function $\mathbf{f}$, that encodes the dynamics of the low-dimensional system, expressing $\mathbf{F}$ in \eqref{eq: dy_eq_full} in the new set of latent coordinates. 
We adopt the framework of Sparse Identification of Nonlinear Dynamics~(SINDy)~\cite{brunton2016discovering}, which assumes that $\mathbf{f}$ can be approximated as a sparse combination of a set of $r$ candidate functions~$\sindyLib(\latentState,\param)\in\R^r$ and coefficients $\sindyCoeff\in~\Rdim^{\latentStateDim\times r}$ resulting in
\begin{align}
    \dot{\latentState} = \mathbf{f}(\latentState, \param) \approx \sindyCoeff\sindyLib(\latentState,\param).
\label{eq: SINDy}
\end{align}
To construct a suitable right-hand side, $r$ potentially useful basis functions, such as polynomial or trigonometric functions are considered, e.g. $\sindyLib(\latentState,\param)=[1, z_1, z_1^2, z_2\beta_1 \dots, \sin(z_1), \beta_1\cos(z_2), \dots]$. The choice of the candidate functions is typically guided by prior knowledge of the physical system and of the parameter/forcing dependency.
The unknown coefficients which determines the active terms from~$\sindyLib$ in~$\mathbf{f}$ are collected in the matrix $\sindyCoeff$ and estimated by sparse regression.
 
VINDy adapts the variational approach behind VAEs to system identification in order to translate SINDy into a probabilistic setting, where UQ of the coefficients $\sindyCoeff$ is directly feasible.
In this setting, each entry $\sindyCoeffentry_{ij}$ is treated as independent scalar variable that determines the contribution of the $i$-th function candidate to the $j-$th equation of system \eqref{eq: SINDy}, defining the dynamics of $z_j$.
With VINDy, we assume the dynamics $\mathbf{f}$, \i.e. $\dot{\latentState}$, to be a random variable generated from $\sindyCoeff$ and $\latentState$ following \eqref{eq: SINDy}\footnote{Without loss of generality, we assume $\param$ as deterministic and consequently drop the dependency for all random variables.}. 
In particular, we assume
\begin{equation}
\dot{\latentState}|\sindyCoeff,\latentState\sim\NormalDist\left(\sindyCoeff\sindyLib(\latentState,\param), \, \tilde{\sigma}^2\bm{I}\right),
\label{eq: dz_dist}
\end{equation}
where $\tilde{\sigma}^2$ is a scalar hyperparameter. Given a prior $p(\sindyCoeff)$ over $\R^{r\times\latentStateDim}$, we aim to  approximate the unknown posterior distribution $p(\sindyCoeff|\dot{\latentState})$ of the model coefficients with~$q(\sindyCoeff|\dot{\latentState})$, searched within a suitable family $\tilde{\mathcal{Q}}$. 
In particular, we use distribution families~$\tilde{\mathcal{Q}}$ that are parameterizable with a set of trainable weight matrices of dimension $r\times\latentStateDim$, that is $\weights_{\sindyCoeff}=
\{{\weights}_{\sindyCoeff_{\vaeMean}},{\weights}_{\sindyCoeff_{\vaeVar}}\}$. 
The entries ${\weightsentry}_{\sindyCoeff_{\vaeMean}}^{ij}, {\weightsentry}_{\sindyCoeff_{\vaeVar}}^{ij}$ directly define the distribution parameters of $\sindyCoeffentry_{ij}$. 
For example, they can represent the mean and variance for a Gaussian distribution $\sindyCoeffentry_{ij}\sim\NormalDist({\weightsentry}_{\sindyCoeff_{\vaeMean}}^{ij},{\weightsentry}_{\sindyCoeff_{\vaeVar}}^{ij})$ or the localization parameter and scale factor in case of a Laplace distribution $\sindyCoeffentry_{ij}\sim~\LaplacianDist({\weightsentry}_{\sindyCoeff_{\vaeMean}}^{ij},{\weightsentry}_{\sindyCoeff_{\vaeVar}}^{ij})$.
In contrast to Gaussian priors, Laplacian priors can act as sparsity-promoting regularization terms within a regression task \cite{Hirsh2022sparsifying}. 
Consequently, they are more prone to provide a parsimonious dynamical model.
More detailed information on the choice of distribution families can be found in~\ref{app: prior}.

Note that the distribution parameters of the approximated posterior for $\sindyCoeff$ do not feature an explicit dependency on the conditioning variable $\dot{\latentState}$. This differs from the posteriors approximated by the encoder and decoder distributions for which their conditioning variables ($\states$ and $\latentState$, respectively) are explicit inputs of the neural network defining their statistical moments (see \eqref{eq: decoder_dist}-\eqref{eq: encoder_dist}). 
To highlight this difference, in the following, we indicate the posterior $\varPosterior(\sindyCoeff|\dot{z};\weights_{\sindyCoeff})$ simply as $\varPosterior(\sindyCoeff)$. In the same way as $\varPosterior(\sindyCoeff)$, we may approximate the posterior on the latent variables $\varPosterior(\latentState)$ with  a distribution parameterized by trainable weights.

Analogously to the objective \eqref{eq: objective1} derived for the VAE, we maximize the log-probability of the latent derivatives $\dot{\latentState}$ in order to find the coefficient distributions explaining the observed dynamics, while reducing the error of approximating the posteriors for $\sindyCoeff$ and $\latentState$, that is
\begin{align}
 \begin{split}
    \log \prob(\dot{\latentState}) 
    &- \KL(\varPosterior(\sindyCoeff) \parallel \prob(\sindyCoeff|\dot{\latentState})) 
    - \KL(\varPosterior(\latentState) \parallel \prob(\latentState|\dot{\latentState}))=
    \\
    &= \expectation_{\sindyCoeff \sim \varPosterior(\sindyCoeff)}
        \left[
            \log(\prob(\dot{\latentState}  \mid \sindyCoeff, \latentState))
        \right] - \KL(\varPosterior(\sindyCoeff) \parallel \prob(\sindyCoeff))
        - \KL(\varPosterior(\latentState) \parallel \prob(\latentState)).
 \end{split}
 \label{eq: VINDy Optimization}
\end{align}
 The derivation of the right-hand side of \eqref{eq: VINDy Optimization}, which translates the theoretical objective in practically computable quantities, is explained in \ref{sec: vindy loss}.

\subsection{Offline training}
VENI and VINDy can be addressed individually by optimizing the objective functions \eqref{eq: objective2} and \eqref{eq: VINDy Optimization}. However these taks are strictly related, as the possibility of identifying parsimonious, accurate dynamical models strongly depends on the choice of reduced coordinates \cite{champion2019data, lusch2018deep}, and the ones provided by VENI are not necessarily optimal, as VAEs do not take dynamics into account. 
This motivates us to tackle VENI and VINDy simultaneously in an unified offline training, which consists in solving
\begin{equation}
    \underset{\weights_{\encoder}, \weights_{\decoder}, \weights_{\sindyCoeff}}{\max  }
    \expectation_{\states\sim \dataset}\left[
     \underbrace{\expectation_{\latentState\sim \varPosterior(\latentState|\states)}[\log\decoder(\latentState)]}_\text{\small Reconstruction} 
     - \underbrace{2\KL(\encoder(\states) \parallel \prior)}_\text{\small Posterior for $\latentState$}
     + \underbrace{\expectation_{\sindyCoeff \sim \varPosterior(\sindyCoeff)}
        \left[
            \log(\prob(\dot{\latentState}  \mid \sindyCoeff, \latentState))
        \right]}_\text{\small Latent dynamics}
    - \underbrace{\KL(\varPosterior(\sindyCoeff) \parallel p(\sindyCoeff))}_\text{\small Posterior for $\sindyCoeff$}
    \right].
     \label{eq: loss}
\end{equation}
The optimization function above represents the mean value of all terms appearing in \eqref{eq: objective2} and \eqref{eq: VINDy Optimization} over the training dataset $\dataset$, where the VINDy latent posterior $\varPosterior(\latentState)$ is replaced by the VAE latent posterior $\encoder(\states)$.

Following this strategy, we simultaneously learn the distribution of the coefficients $\sindyCoeff$, together with the encoder $\encoder$ and decoder $\decoder$ distributions. 
We optimize all the unknown coefficients $\weights_{\encoder}, \weights_{\decoder}, \weights_{\sindyCoeff}$ via backpropagation, using standard gradient-based methods as, e.g., ADAM \cite{adam}.
The only arrangement we have to take is the standard \textit{reparameterization trick} in VAEs \cite{kingma2013auto, rezende2014stochastic} to allow backpropagation of \eqref{eq: loss} through the networks. 
Indeed, sampling $\latentState$ directly from the distribution defined by the encoder is not a differentiable operation, thus has no gradient. 
A remedy can be found if the family of distributions for the encoder is closed under linear transformations. By introducing a noise term, sampled from the same family of distributions, as an artificial input to the network, the sampling process is rewritten in such a way that the randomness is decoupled from the learnable parameters. In case of a Gaussian distribution, the noise term acknowledges~$\vaeNoise\sim \NormalDist (\bm{0},\bm{I})$ and a sample from the encoder posterior can be drawn according to 
\begin{equation}
\latentState^{(i)}=\hat{\latentState}(\states^{(i)}) + \vaeNoise * \vaeVar_{\encoder}(\states^{(i)}),
\label{eq: reparam}
\end{equation} 
where $*$ indicates the element-wise product. 
This reparameterization trick also holds, e.g. for Laplace distributions in which case $\vaeNoise \sim \LaplacianDist (\bm{0},\bm{I})$.
From \eqref{eq: reparam}, it follows that a sample for the latent state derivatives can be obtained by applying chain rule to the full-state derivatives
\begin{equation}
\dot{\latentState}^{(i)}
= \nabla_{\states} \hat{\latentState}(\states^{(i)})\dot{\states}^{(i)} + \vaeNoise * \nabla_{\states} \vaeVar_{\encoder}(\states^{(i)})\dot{\states}^{(i)}.
\label{eq: reparam_dz}
\end{equation}
In the following section, we present how to translate the general probabilistic objective \eqref{eq: loss} into a data-driven optimization problem.

\subsubsection{Data-driven loss}
Thanks to the assumptions and choices on the family of distributions (see, e.g., \eqref{eq: decoder_dist} and \eqref{eq: dz_dist}), the objective \eqref{eq: loss} can be translated into the following practical, data-driven optimization problem, where the expectation is approximated by the sample mean over the training set:
\begin{align}
\begin{split}
    \underset{\weights_{\encoder}, \weights_{\decoder}, \weights_{\sindyCoeff}}{\min  } \biggl( \frac{1}{n_s}\sum_{i=1}^{n_s} 
    &\underbrace{\lambda_1 \norm{\states^{(i)} - \hat{\states}^{{(i)}} }_2^2}_\text{\small Reconstruction}
    + \underbrace{\lambda_2\KL \left(\encoder(\states^{(i)}) \parallel \prior\right)}_\text{\small Posterior for $\latentState$}
    + \underbrace{\lambda_3{\norm{\dot{\latentState}^{(i)} - \sindyCoeff^{(i)}\sindyLib(\latentState^{(i)},\param)}}_2^2}_\text{\small Latent dynamics}  \\
     + &  \underbrace{\lambda_4\KL \left( \varPosterior(\sindyCoeff) \parallel p(\sindyCoeff)\right)}_\text{\small Posterior for $\sindyCoeff$}
    + \underbrace{\lambda_5 \norm{\dot{\states}^{(i)} - \nabla_{\latentState}\hat{\states}^{{(i)}} \sindyCoeff^{(i)}\sindyLib(\latentState^{(i)},\param)}_2^2}_\text{\small Full dynamics}
    \, \biggr).
    \label{eq: loss_data}
\end{split}
\end{align}

Here,~$\states^{(i)}$ and~$\dot{\states}^{(i)}$ indicate the full state and time derivatives of the $i-$th training snapshot; $\latentState^{(i)}$, $\dot{\latentState}^{(i)}$ are the latent counterparts; 
while $\sindyCoeff^{(i)}$ is a sample drawn from the coefficient's approximate posterior $q(\sindyCoeff)$. 
Finally, $\hat{\states}^{(i)}$ indicates the mean of the decoder output distribution generated from a latent sample~$\latentState^{(i)}$.\\
The first term (reconstruction) of \eqref{eq: loss_data} ensures that a full state can be recovered from the latent variables; the second one (posterior for $\latentState$) pushes the approximated posterior distribution of the latent variables represented by the encoder~$\encoder(\states)$ closer to the selected prior~$\prior$; the third one (latent dynamics) is responsible for ensuring that the identified dynamical system matches the data; the fourth one (posterior for~$\sindyCoeff$) compels the approximated posterior distribution of the VINDy coefficients to be proximate to the selected prior~$p(\sindyCoeff)$; and the last one (full dynamics) acts as a regularization term that accounts for consistency between the reference time derivatives of the data $\dot{\states}^{(i)}$ and the mean value of the reconstruction from the approximated latent derivatives, as suggested in \cite{champion2019data}. 
The weighting coefficients $\{\lambda_i\}_{i=1}^{5}\in \R^+$ are hyperparameters that determine the contribution of each term. 
For general guidance on tuning these hyperparameters, as well as details on the loss derivation and its constituent terms, we refer the reader to \ref{app: solve_veni_vindy}.

\subsubsection{Sparsity promotion by PDF thresholding}
\label{sec: pdf thresholding}
Following the joint optimization of VENI and VINDy in the offline training, a post-processing step can be employed to promote sparsity. In detail, coefficients with large uncertainty, which do not have a clear and consistent contribution to the dynamics, are pruned. 
The basic idea is to discard those coefficients whose identified probability density function at zero exceeds a specified threshold, i.e.~$\text{pdf}_\LaplacianDist(0) > \pdfThreshold$.
The threshold value, denoted as~$\pdfThreshold$, functions as a hyperparameter, facilitating the adjustment of sparsity within the identified system. As a further refinement, the model could be fine-tuned on the retained coefficients to obtain more accurate estimates of their probability density functions.

By basing the selection on the probability density function of the coefficients rather than their raw magnitude, this strategy provides a more grounded criterion for sparsity promotion and mitigates the risk of discarding small but consistently significant terms.
This is in contrast to the sequential thresholding method, as outlined in~\cite{brunton2016discovering}, where all coefficients of small magnitude are eliminated. 
As a possible extension, the sparsity threshold could be treated as a learnable parameter during training, removing the need for manual tuning; however, this might result in a more complex optimization problem.

\subsection{Variational Inference with Credibility Intervals (VICI)}
Once the offline training phase is complete, the fitted model can be queried online to generate new solutions to \eqref{eq: dy_eq_full} for unseen initial conditions~$\states_0$ and input parameters~$\param$. Specifically, we employ the online VICI procedure to simulate the evolution of  system states and generate uncertainty intervals for the estimated solution trajectories, as illustrated in Fig.~\ref{fig: vici}. 

For a given initial condition~$\states_{0}$ in the full space, we sample~$m$ latent initial conditions from the approximated posteriors, \i.e.~$\{\latentState_0^{(i)}\}_{i=1}^{m}\stackrel{\text{iid}}{\sim}\encoder(\states_0)$ and coefficients~$\{\sindyCoeff^{(i)}\}_{i=1}^{m}\stackrel{\text{iid}}{\sim}\varPosterior(\sindyCoeff)$.
This differs from the standard generative methods of VAE, where the encoder is usually ignored during testing, and new solutions are generated directly from the latent variables sampled from the prior. 
In our framework, we are taking into account time and dynamics, therefore it is essential to employ the encoder to generate physics-consistent initial conditions in the latent space.  \\
Then, for each pair~$(\latentState_0^{(i)}, \sindyCoeff^{(i)})$, we use standard time-stepping schemes, such as Runge-Kutta methods, to evolve the system of ordinary differential equations~$\dot{\latentState} = \sindyCoeff^{(i)}\sindyLib(\latentState,\param)$ from the initial condition~$\latentState_{0}^{(i)}$ over a set of discrete time-steps $\mathcal{T}=\{0,t_1, \ldots, T\}$. Finally, the so-computed latent trajectories~$\{\latentState^{(i)}(t): t\in\mathcal{T}\}_{i=1}^{m}$ are passed to the decoder that outputs the mean of the full state solution trajectories~$\{\hat{\states}^{(i)}(t)=\hat{\states}(\latentState^{(i)}(t)): t\in\mathcal{T}\}_{i=1}^{m}$. Note that we avoid sampling at each time step~$t$ from the corresponding posterior~$\states^{(i)}(t)\sim~\decoder(\latentState^{(i)}(t))$, in order to preserve in the full space (possible) regularity properties in time of the latent trajectories. Uncertainty bounds for the approximated solution are then calculated from the statistical moments of the set of~$m$ trajectories.
\begin{figure}[t!]
    \centering
    \input{Figures/vici/vici}
    \caption{Schematic representation of the online VICI procedure to generate new solutions for a given initial condition $\states_0$ and set of parameters $\param$. We first sample multiple instances of the corresponding latent initial conditions and coefficients of the dynamical model, each defining an ODE system. Then each dynamical system is integrated in time through standard time-stepping schemes, resulting in multiple latent trajectories. These latter are finally processed by the decoder mean to obtain full state trajectories. Predictions and the corresponding UQ are computed directly from the statistical properties of the approximated solution trajectories.} 
    \label{fig: vici}
\end{figure}

\section{Results}
\label{sec: results}

\subsection{R\"ossler System: Identification of a low-dimensional chaotic systems under varying noise-levels}

\noindent In order to assess the potential of our novel VINDy approach alone (without autoencoder), we present results for a low-dimensional chaotic system. Therefore, we consider the R\"ossler system
\begin{align}
    \begin{split}
        \dot{\roeA}  &= -\roeB -\roeC, \\
        \dot{\roeB} &= \roeA + \roeCoeffA \roeB,\\
        \dot{\roeC} &= \roeCoeffB + \roeC(\roeA - \roeCoeffC),
    \end{split}
    \label{eq: roessler}
\end{align}
where~$\latentState=[\roeA, \roeB, \roeC]^\transpose$ is the vector of the system states and~$\modelParams=[\roeCoeffA=0.2, \roeCoeffB=0.2, \roeCoeffC=5.7]^\transpose$ contains the system parameters. Model hyperparameters and dataset configurations are summarized in the \ref{sec: network_implementation} in \cref{tab:network hyperparameter}-\ref{tab:data hyperparameter}.

\begin{figure}
    \centering
    \includegraphics[]{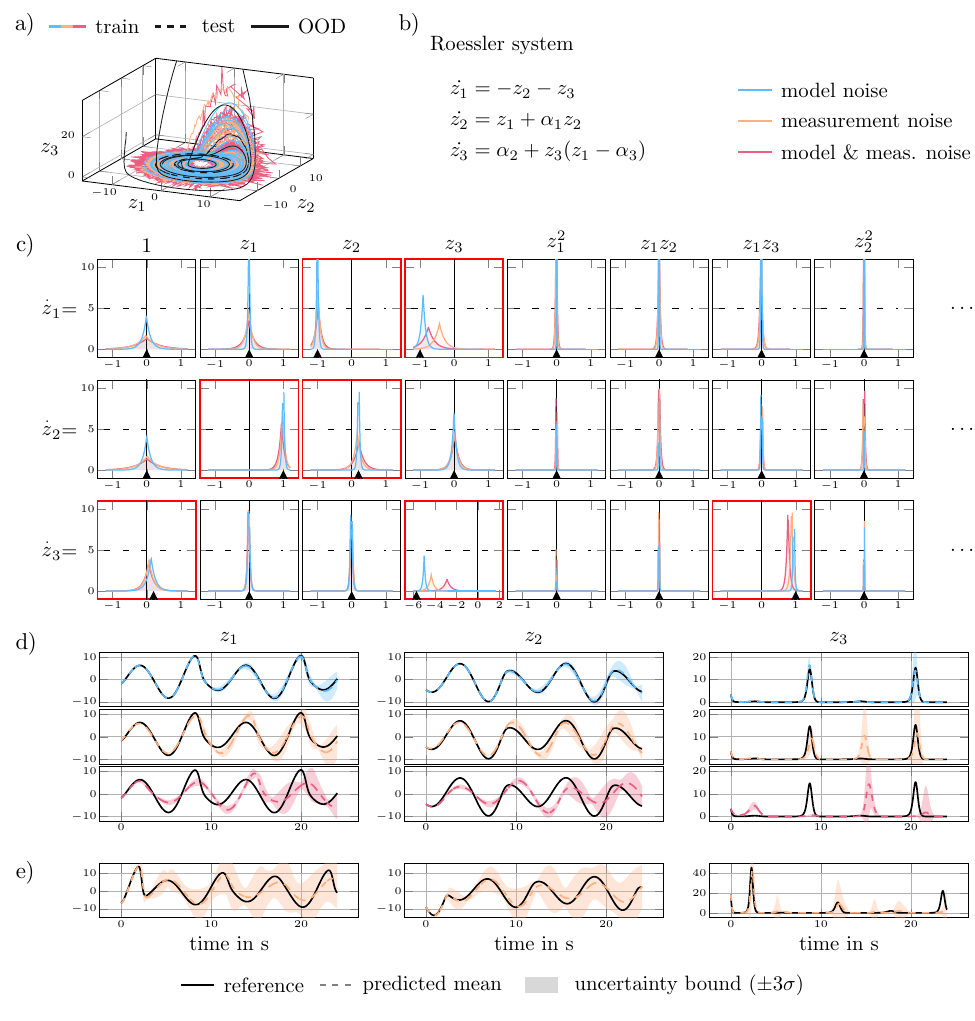}
    \caption{Approximation results for the R\"ossler system. Different types of noisy data (each represented by a different color) a) are generated by the R\"ossler system, b) serve as basis for identifying the distributions of coefficients, and c) are compared against the true values (indicated by triangles at the bottom of each axis). Coefficients appearing in the original equations are emphasized with bold red axes, while the PDF threshold is marked by a dashed line. The resulting temporal evolutions of the identified systems, including UQ, are shown in d) for an initial condition (IC) within the training regime and in e) for an IC outside the training domain (out-of-distribution inference).}
    \label{fig: roessler}
\end{figure}

\paragraph{Data generation} The data used to train our model consists of 30 simulations of the R\"ossler System  \eqref{eq: roessler} with normal randomly distributed initial conditions~$\latentState(t_0)\sim\NormalDist([-5,-5,0]^\transpose, \standardDeviation^2 \bm{I})$ with $\standardDeviation=2.25$. 
Each simulation has a time horizon of~$T=24$ and consists out of $N_t^\text{train}=2000$ time steps.
We consider two sources of uncertainty: measurement noise and model noise. For the former we apply multiplicative noise to the state measurements, \i.e.~$\latentState_{\text{noise}}(t)=\noise*\latentState(t)$ with~$\noise\sim\LaplacianDist\NormalDist(\bm{0},\measNoiseFactor\bm{I})$ drawn from a log-normal distribution individually for every sample. The time derivatives are numerically computed from those noisy samples. The model uncertainty is realized as noise in the model parameters which are drawn from a normal distribution~$\modelParams\sim\NormalDist\left([\roeCoeffA, \roeCoeffB, \roeCoeffC]^\transpose, \text{diag}([\standardDeviation_{\roeCoeffA}^2, \standardDeviation_{\roeCoeffB}^2, \standardDeviation_{\roeCoeffC}^2]^\transpose)\right)$ centered around the correct values with standard deviations~$\standardDeviation_{\roeCoeffA}=\modelNoiseFactor{\roeCoeffA}$,~$\standardDeviation_{\roeCoeffB}=\modelNoiseFactor{\roeCoeffB}$, and~$\standardDeviation_{\roeCoeffC} =\modelNoiseFactor{\roeCoeffC}$. 
For the experiments in \cref{fig: roessler}, we applied~$\modelNoiseFactor=0.1$ for model noise and~$\measNoiseFactor=0.05$ respectively, while~$\modelNoiseFactor=0.2$ and~$\measNoiseFactor=0.1$ are used for the combined noise scenario.
The test data consist of simulations that differ from the training data in their initial conditions, while using the correct parameter values. 
We consider two testing scenarios: initial conditions (ICs) sampled within the training regime, and out-of-distribution (OOD) ICs, drawn from a Gaussian distribution
$
\latentState(t_0)\sim \NormalDist\!\left([-10,\,-10,\,20]^\transpose, \,\standardDeviation^2 \bm{I}\right).
$

\paragraph*{Model}
For the VINDy layer, we use a polynomial library of second degree, including bias and interactions, for a total of $r=10$ candidate features. We consider Laplacian priors~$\sindyCoeff_{ij}\sim\LaplacianDist(0,1)$ for the corresponding coefficients. The model is trained for 500 epochs. At the end of the training, all coefficients~$\sindyCoeff_{ij}$ with a probability density function above a threshold of~$pdf_{\sindyCoeff_{ij}}(0)>5$ are removed.

\paragraph{Results} We identify governing equations using VINDy with a polynomial library of candidate functions under model and measurement noise. 
The impact of the respective noise on the training data can be seen in \cref{fig: roessler}. 
The resulting trajectories under model noise closely match the ground truth, and a similar agreement is observed for measurement noise. 
Even in the OOD case, the predicted trajectories remain well aligned, albeit with broader uncertainty intervals. 
The approach only begins to reach its limits when high model and measurement noise are present simultaneously. 
However, even in the latter case, the method identifies all the essential terms that originally appear in the governing equations, regardless of the type of uncertainty hampering the data.
At the same time, the irrelevant coefficients are centered around zero with corresponding high values of their PDF. For moderate noise levels, the resulting coefficient distributions center around the reference value with high confidence, whereas higher noise levels lead to more conservative estimates and wider distributions. 

\subsection{Identification of a reaction-diffusion problem}
\label{sec: RD}
\noindent We consider a reaction-diffusion system governed by
\begin{equation}
    \begin{aligned}
&\dot{u}=\left(1-\left(u^{2}+v^{2}\right)\right) u+\mu\left(u^{2}+v^{2}\right) v+d_1\left(u_{x x}+u_{y y}\right),\\ 
&\dot{v}=-\mu\left(u^{2}+v^{2}\right) u+\left(1-\left(u^{2}+v^{2}\right)\right) v+d_2\left(v_{x x}+v_{y y}\right) , 
\end{aligned}
\label{eq: RD_eqs}
\end{equation}
where $\mu = 1.0$ and $d_1 = d_2 = 0.01$ are coefficients that respectively regulate the reaction and diffusion behaviors of the system. The problem is defined over a spatial domain $[-L,L]^2$ for $L=10$ and a time span $t\in [0,T]$ for $T=40$,
periodic boundary conditions are prescribed, and the initial condition is defined as
\begin{equation}
  \begin{aligned}
     u(x,y,0;\beta) = v(x,y,0;\beta) = \tanh{\left(\beta\sqrt{x^2 + y^2}\cos{\left((x+iy)-\beta\sqrt{x^2 + y^2}\right)}\right)}\, , 
  \end{aligned}
\label{eq: RD_ic}
\end{equation}
depending on the parameter $\beta \in [0.7, 1.1]$.
System \eqref{eq: RD_eqs} generates spiral waves, which represent an attracting limit cycle in the state space \cite{floryan2022data} whose radius is modulated by the parameter $\beta$.
Our goal is to create a generative model to compute the entire space–time solution for a new instance of the parameter $\beta$. 
\ref{sec: network_implementation} summarizes the model hyperparameters and data configuration parameters in \cref{tab:network hyperparameter}-\ref{tab:data hyperparameter}.

\paragraph{Data generation}
\noindent A grid of $N_\beta = 20$ values is considered for the parameter $\beta$ equispaced with $[0.7, 1.1]$. The parameter values are randomly partitioned into training and testing subsets such that $N_\beta^\text{train} = 16$ and $N_\beta^\text{test} = 4$. For each choice of $\beta$, numerical solutions for $u$ and $v$ are calculated by solving the PDE system \eqref{eq: RD_eqs} with initial condition~\eqref{eq: RD_ic} by using the Fourier spectral method \cite{trefethen2000spectral} with time step~$\Delta t = 0.05$ on an equispaced spatial grid with spatial step~$\Delta h =  0.4$.
For the training set, the solutions are calculated for a limited time window to $T_\text{train}=20$, while the testing data reach the final time $T=40$. In addition, training data are corrupted by $20\%$ log-normal multiplicative noise to simulate measurement noise.
Training and testing snapshots are stacked in matrices $\dataset_\text{train}\in\mathbb{R}^{N_t^\text{train} N_\beta^\text{train} \times N}$ and $\dataset_\text{test}\in\mathbb{R}^{N_t^\text{test} N_\beta^\text{test}\times N}$, where $N_t^\text{train} = 400$ and $N_t^\text{test} = 800$ are the number of samples for training and testing sets respectively, and $N = 2500$ is the number of spatial degrees of freedom. We preliminary reduce system dimension to $N_\text{POD}=32$ by projecting both training and testing data onto the reduced basis obtained using POD on $\mathbf{X}_\text{train}$. 

\paragraph{Model}
\noindent Regarding the VENI structure, the encoder consists out of a feed-forward neural network with 3 hidden dense layers of 32, 16 and 8 units, while the decoder network has a symmetrical structure.
Spiral waves generated by the system \eqref{eq: RD_eqs} can be approximately captured by two oscillating spatial modes \cite{champion2019data}, thus setting the number of latent variables $n=2$.
For VINDy, we employ a library of polynomial functions of the latent variables $\vb{z} = \left[z_1, z_2\right]^\top$ up to the third degree with interaction and bias. No parametric dependency is included in the library, as parameter $\beta$ affects only the initial conditions. Regarding the choice of prior distributions on the latent variables and on the VINDy coefficients, we consider Gaussian and Laplacian distributions, respectively, namely $\latentState\sim\NormalDist(\vb{0}, \bm{I})$ and $\Xi_{ij}\sim\LaplacianDist(0, 1)$ for each entry.

\begin{figure}[t!]
    \centering
    \includegraphics[width=\textwidth]{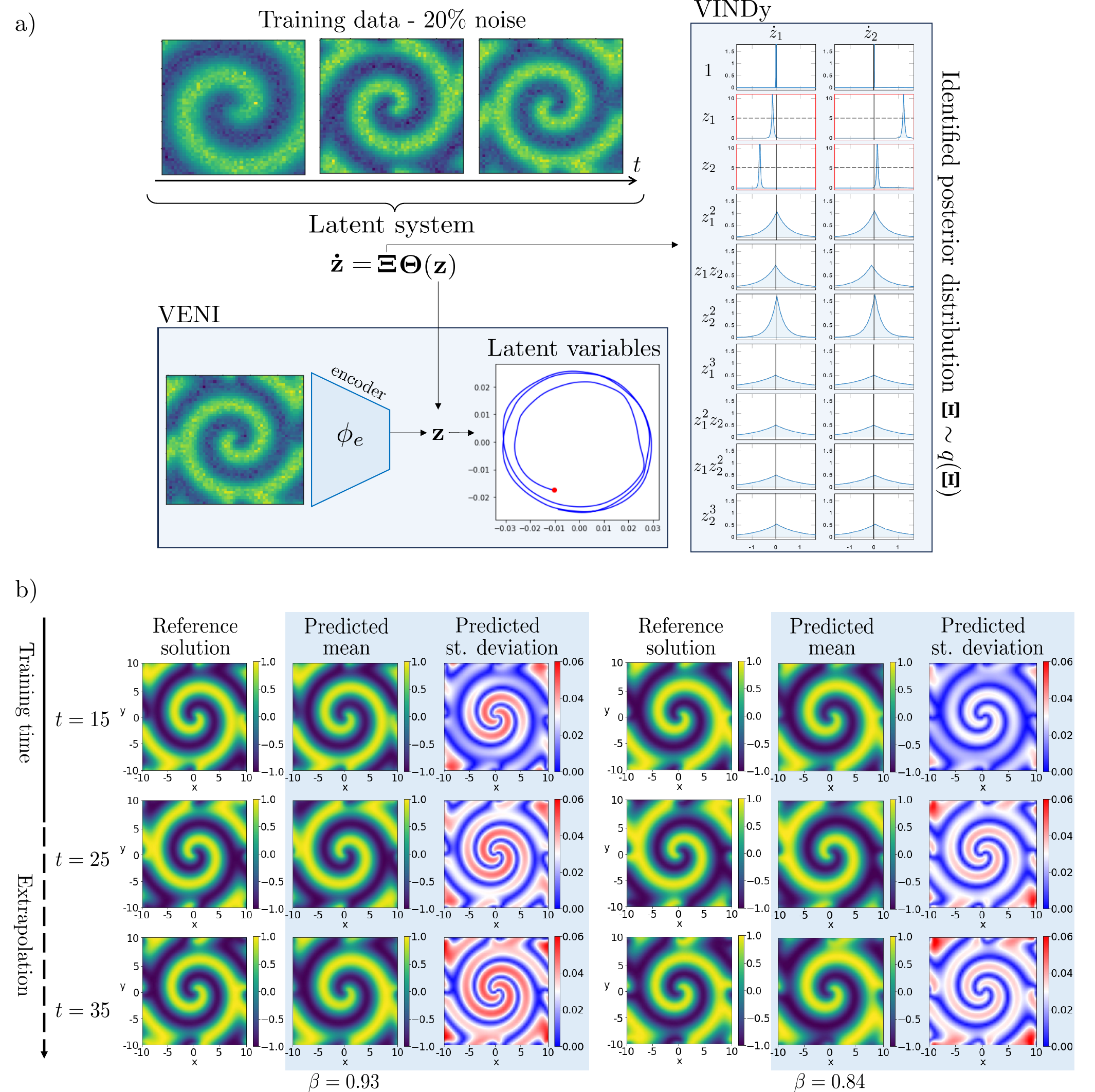}
    \caption{
    Results of the offline training a), consisting of the VENI and VINDy steps, for the reaction-diffusion problem. In the VINDy box, the identified posterior distributions of the coefficients are displayed. We note that the mixed linear terms, which are the dominant terms for the observed dynamics, are accurately identified as significant. In the VENI box, we highlight how the method is capable of learning an encoding function from noisy data, resulting in two latent variables that exhibit the expected oscillatory behavior. 
    A comparison of solution fields between the approximation and the numerical reference solution is given in b) for a training $t=15$ and extrapolated time instances $t \in \{20,35\}$, while forward uncertainty quantification at the different reduction levels, performed with the VICI procedure.} 
    \label{fig: RD}
\end{figure}

\paragraph{Results} The posterior distributions identified by our method for the model coefficients are depicted in Fig.~\ref{fig: RD}. 
We note that the mixed linear terms, which are the dominant terms for the observed oscillatory dynamics, are accurately identified as significant with high confidence. 
In addition, our approach promotes sparsity in the dynamics, since most nonlinear terms are represented by distributions with high probability density function values at zero.
Fig.~\ref{fig: RD} also presents the phase space representation in the identified latent coordinates, showcasing the oscillatory pattern and the attracting limit cycle. This demonstrates our method's ability to preserve interpretable insights into the latent space.

To assess the accuracy and reliability of VICI, we evaluate its performance for initial conditions and test parameter instances unseen during training. We predict the approximate solutions up to a final time of $T=40$, while we only use training data up to $T_\text{train}=20$. In Fig.~\ref{fig: RD}, we illustrate the mean and standard deviation for over $m=100$ realizations of the predicted solutions of the entire field, and compare them with the noise-free test data. 
We note how the method is able to accurately detect the spatial dynamics, even when extrapolating on a time window twice longer than the training range. 

\subsection{MEMS resonator: Identification of structural mechanical second-order systems}
\label{sec: results_MEMS}

\noindent We consider a straight beam MEMS resonator, excited at resonance, represented as double clamped beam as illustrated in \cref{fig: beam}. 
The considered beam MEMS device has length $L=~1000$\micr with a rectangular cross-section of dimensions 10\micr$\times$24\micr, made of isotropic polysilicon \cite{corigliano2004mechanical}, with density $\rho=~2330$\,Kg/m$^3$, Young modulus $E=167$\,GPa and  Poisson coefficient $\nu=0.22$ {(see Fig.~\ref{fig: beam}a). 
The governing PDEs for this solid mechanics problem in large transformations \cite{malvern1969introduction} are formulated in terms of the displacement field $\bfu$ as follows:
\begin{subequations} 
	\label{eq:strongmech}
	\begin{align}
		\rho_0 \ddot{\bfu}({\varkappa},t) - \nabla\cdot\bfP(\varkappa,t)- 
		\rho_0 \bfB(\varkappa,t;\param)=\bfzero, \; 
		&   \quad (\varkappa,t)\in \Omega_0 \times (0,T),
		\label{eq:strongmech_a}
		\\
		\bfP(\varkappa,t)\cdot\bfN(\varkappa)= \bfzero,\; 
		& \quad (\varkappa,t)\in \partial \Omega_N \times (0,T),
		\label{eq:strongmech_c}
		\\
		\bfu(\varkappa,t)=\bfzero,\; 
		& \quad (\varkappa,t)\in \partial \Omega_D \times (0,T), 
		\label{eq:strongmech_d}
	\end{align}
\label{eq:PDE_beam}
\end{subequations}
where $\Omega_0$ is the domain occupied by the device in the undeformed configuration, described by material coordinates $\varkappa$, and $T=1755\,\mu\text{s}$. Eq.~\eqref{eq:strongmech_a} expresses the conservation of momentum, where $\rho_0$ is the initial density, $\bfP$ is the first Piola-Kirchhoff stress and $\bfB$ are the external body forces. In this application, we consider a periodic body load proportional to the first vibrational eigenfunction, with the amplitude and frequency given by the parameters $F$ and $\omega$, respectively. The input parameter vector hence becomes $\param~=[\omega, F]^\transpose$, defined over a closed and bounded set of dimension two. Eq.~\eqref{eq:strongmech_c} and \eqref{eq:strongmech_d} define the 
homogeneous boundary conditions on the Neumann  and Dirichlet and boundaries denoted as $\partial\Omega_N$ and  $\partial\Omega_D$, respectively, and $\bfN$ denotes the outward-directed unit normal on the boundary of $\Omega$. For more details regarding the application, we refer to~\cite{conti2023reduced}.
A comprehensive summary of network hyperparameters and dataset settings can be found in \cref{tab:network hyperparameter}-\ref{tab:data hyperparameter} in the \ref{sec: network_implementation}.

\paragraph{Data generation}
We consider $N_\beta = 56$ parameter instances, corresponding to 28 values for the frequency $\omega$, selected in the range $[0.526, 0.564]\,\text{rad}/\mu\text{s}$ with a finer sampling around the natural frequency $\omega_0$, and for two values of the forcing amplitude $F\in\{0.125,0.250\}\,\mu\text{N}$. In total, $N_\beta^\text{train}=28$ parameter instances are randomly selected to construct the training set and the remaining $N_\beta^\text{test}=28$ are retained for testing. Numerical solutions for the displacement are computed for each parameter configuration, by discretizing the system~\eqref{eq:PDE_beam} using the finite element method over a spatial mesh consisting of $N_\text{dof}=7821$ degrees of freedom as in \cite{conti2023reduced}. 
Training data are calculated up to $T_\text{train} = 1091\,{\mu}s < T$ and corrupted by additive noise following a normal distribution whose scale is proportional to the mean displacement level in all training simulations. 

As we are modeling the beam as a second-order system, we require to have access or compute the second time derivatives $\ddot{\states}$ for training in addition to the system states $\states$ and their time derivatives $\dot{\states}$. 
This step is particularly challenging for second-order systems, where even low noise increases dramatically on the acceleration level as it grows with each numerical derivation.
For the beam data, the ratio between the maximum amplitude occurring in the noisy training data and the noise-free reference data on acceleration level across all simulations is~$\numprint{4.52455}\pm \numprint{2.26626}$.

 As in the previous example, the data (including displacements, velocities, and accelerations) are preliminary projected onto a POD basis of $N = N_\text{POD} = 3$ modes (extracted from the noisy displacement training data). 

\paragraph{Model}
The encoder is composed of three hidden layers that use 32 neurons each. The decoder has symmetrical structure. Although the dynamics of this system lies on a two-dimensional manifold in the phase space, we manage to reduce the dimensionality to a single latent variable $n=1$, as the dependence on the velocity is the minimal for the master bending mode that we inspect herein. As in \eqref{eq:PDE_beam}, structural mechanics problems are often governed by second-order governing equations. Therefore we model the reduced dynamics with a second-order ODE,  by augmenting the reduce states with the first derivatives $\dot{z}$:
\begin{equation}
    \frac{\text{d}}{\text{dt}}
    \begin{bmatrix}
        \;z\; \\
        \;\dot{z}\;
    \end{bmatrix}
    =
    \mathbf{f}\left(z, \dot{z};\param 
    \right)
    \approx
    \begin{bmatrix}
        \;\dot{z}\; \\  \;\sindyCoeff\sindyLib(z,\dot{z};\param)\;
    \end{bmatrix}.
\end{equation}
The corresponding accelerations at latent level~$\ddot{\latentState}$ can be derived using chain rule as it is done for the velocities~$\dot{\latentState}$. 
%
Note that in case we did not want to exploit any prior knowledge on the second-order structure of the problem, we could have equivalently modeled    the problem by considering $n=2$ latent variables and identifying a first-order ODE system as in the previous example. The library $\sindyLib$ features polynomials up to the third degree with respect to both $\latentState$ and $\dot{\latentState}$. Moreover, since the beam resonator is forced harmonically with amplitude and frequency parameterized by $F$ and $\omega$, we incorporate the parameter dependency by including the library term $F \cos(\omega t)$. We set Gaussian priors on the reduced states $[\latentState, \dot{\latentState}]^\top\sim\NormalDist(\vb{0}, \bm{I})$ and Laplacian prior on each of the VINDy coefficient distributions $\sindyCoeff_{ij}\sim\LaplacianDist(0,1)$. Training is performed for 2500 epochs, and the weights of the networks that performed best in terms of total loss on a validation subset of the training data used for inference. 
After training we apply PDF zero tresholding to only keep the terms that are important to describe the dynamics. In particular, all terms with a $pdf_{\sindyCoeff_{ij}}(0)>1.9$ are neglected in the model.

\paragraph{Results} The clamped-clamped beam is a classical structure showing Duffing-like hardening behavior \cite{corigliano2018mechanics,FRANGI2019} that can be accurately described by the following simple model:
\begin{equation}
    \ddot{z} = -\omega_0^2z - 2\xi\omega_0\dot{z} - \gamma z^3 - \alpha F \cos(\omega t),
    \label{eq: beam_NF}
\end{equation}
where~$-\omega_0^2$ represents the natural frequency and~$-2\xi\omega_0$ the damping of the device.
Moreover, the cubic term, with $\gamma>0$, accounts for the hardening response and the absence of quadratic terms should be remarked.
\begin{figure}[t]
    \centering
    \includegraphics{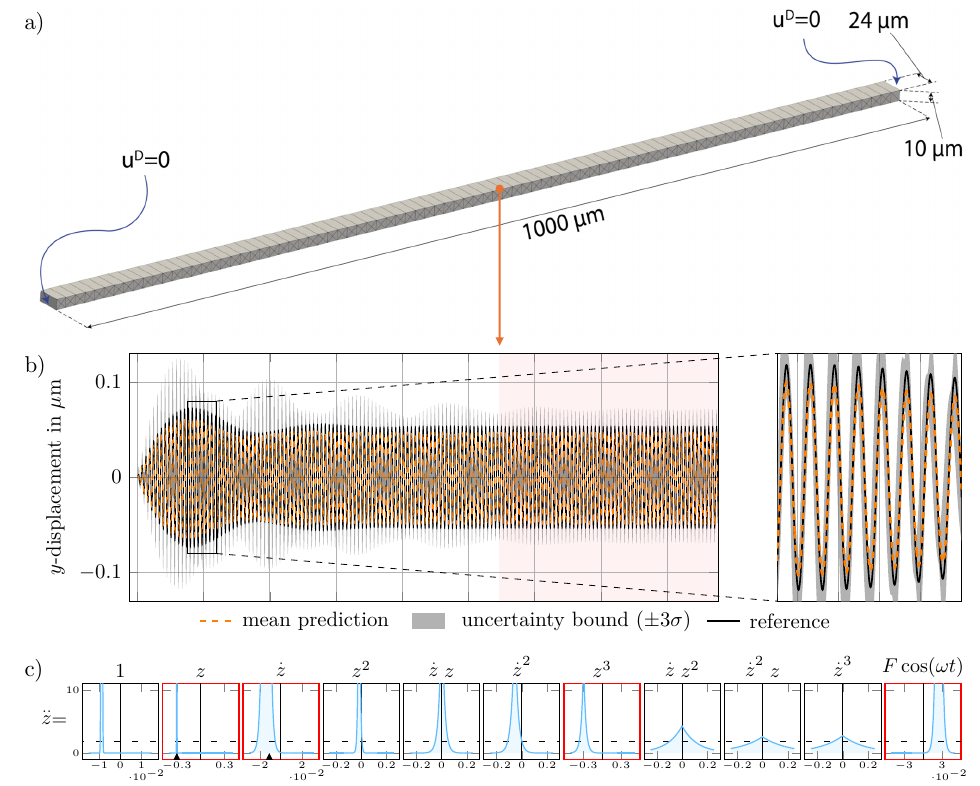}
    \caption{Schematic representation of the beam MEMS resonator a) with the mesh used in the FOM simulations. 
    The physical reconstruction of latent mean trajectories and corresponding uncertainties for a node in the middle of the beam where the most dynamics occur are given in b). For training, only the first $T_\text{train}=1091\,\mu$s are used; the remaining time for which the model extrapolates is indicated with a red background. The identified coefficient distribution c) show that the terms appearing in the normal form of the observed high-dimensional system (framed in bold red, correct values that result from the natural frequency and the damping of the system are marked with a black triangle at the bottom of the according axis, the PDF threshold value is indicated as dashed line) are correctly identified by our method.}
    \label{fig: beam}
\end{figure}
As the objective of VINDy is to identify a sparse reduced system, we wish to unveil the underlying normal form \eqref{eq: beam_NF} of the observed high-dimensional system, directly from data. 
For the considered example, we can see from \cref{fig: beam} that the terms indicated as relevant by our method correspond to those present in the normal form plus an additional bias term of small magnitude. 
Our method effectively and accurately identifies the actual coefficients of the linear terms associated with the natural frequency and the damping, see Fig.\ref{fig: beam}.
This is achieved without requiring access to or integration of knowledge from the FOM, relying solely on noisy data.
In addition, the \textit{full dynamics} term in the loss function \eqref{eq: loss_data} promotes consistency between the latent dynamics and the decoded trajectories, allowing key nonlinear features of the MEMS system -- such as the hardening response, loss of uniqueness, and jump phenomena -- to be preserved after decoding~\cite{conti2023reduced}.
\cref{fig: beam} also shows that the mean full-time evolutions produced by the VICI procedure for new parameter configurations follow the reference, while the standard deviations capture the discrepancies.

To assess the performance of the proposed method, we provide in \ref{sec: comparison} a benchmark comparison against state-of-the-art approaches, including Fourier neural operators \cite{li2020fourier} and Gaussian Process-based ROMs, highlighting the unique features of our framework and main advantages in terms of accuracy, computational efficiency, scalability, uncertainty quantification, extrapolation capabilities and physical interpretability.

\section{Discussion and final remarks}
\label{sec: conclusions}
The results demonstrate the effectiveness of the proposed VENI, VINDy, VICI framework in constructing accurate, interpretable, and uncertainty-aware ROMs directly from time-series data. 
This is achieved by integrating probabilistic modeling with data-driven discovery in a unified \textit{physical generative} framework.
The identified ROMs are consistent with the underlying physics, preserving computational efficiency while providing robust UQ, even in situations with significant noise or limited data. 
VINDy’s capability to select essential terms in the dynamics allows for the discovery of governing equations that offer insights into the fundamental mechanisms of complex systems. 
In \ref{sec: comparison}, we present a benchmark comparison that highlights the unique characteristics of the proposed framework and its advantages relative to alternative state-of-the-art modeling approaches, including Fourier neural operators \cite{li2020fourier} and Gaussian process-based ROMs.

By modeling both state variables and model coefficients as probability distributions, our framework accommodates measurement noise and model uncertainty, enhancing the robustness and reliability of the ROM's predictions.
Several advantages are noted and exploited.
First, the estimated distributions of model coefficients provide valuable insights into the uncertainties in function terms, guiding model selection and refinement. 
Second, sparsity-promoting priors, such as Laplacian priors, in combination with PDF thresholding provide parsimonious models \cite{Hirsh2022sparsifying}.
Sparsity is fundamental for interpretability and generalization, as it enables the identification of essential terms in the dynamics, thereby revealing the governing equations underlying the observed system among an infinite range of possible models.
For example, in the Rössler system, our method accurately identifies the governing equations by estimating the essential terms and their coefficients.
In the reaction-diffusion system, the primary dynamics is effectively modeled as a linear oscillator, with the phase space in the latent coordinates accurately approximating the attracting limit cycle.
In the MEMS resonator application, the framework autonomously discovers the governing dynamics, unveiling the \textit{normal form} of the system. 
This representation not only captures the dynamics in the sparsest form but also provides direct insights into the linear coefficients (e.g., natural frequency and damping) and nonlinear terms determining the system’s dynamics.

Unlike traditional methods that enforce sparsity based solely on coefficient magnitude, the proposed framework leverages probability density functions of the coefficients. 
This allows coefficients with small magnitudes but substantial impacts on the system's dynamics to be retained.
For example, in the beam case, the forcing term -- despite its small amplitude -- is essential for accurately capturing the system’s non-autonomous behavior. 
While magnitude-based approaches might discard this term, our method retains it, highlighting its importance.

In addition to the UQ directly provided by the approximated posterior on latent states and dynamical models, the framework enables UQ on physical solution trajectories using the ensemble strategy of VINDy. 
The generative nature of the method allows the creation of an ensemble of dynamical models through sampling, at virtually no additional cost, unlike standard ensemble-based UQ methods that require training of multiple instances of a model.
By integrating these models over time and mapping the resulting ensemble of latent trajectories to physical space, we gain direct insights into how uncertainty in the spatial solutions evolves over time.

While uncertainties are currently modeled globally at the system level -- reflecting the assumption of a single set of time-invariant governing equations with parametric dependence -- the framework admits a natural extension to component-wise structure by assigning dedicated latent variables and corresponding governing equations to distinct subsystems, thereby localizing uncertainty across system partitions when warranted by the application.

As with most deep learning approaches, the performance of our method can be sensitive to hyperparameters, including the weighting of loss terms, the choice of priors, and the VINDy library of candidate functions. 
Another limitation is that, while the approach effectively forecasts solution trajectories via the identified latent dynamical model, the decoder does not guarantee strong generalization to spatial patterns entirely outside the training data, which may result in unphysical reconstructions in extrapolation scenarios. 
This limitation could be alleviated by employing more advanced architectures as building blocks for the VAE; for instance, graph neural networks \cite{zhou2020graph} may improve spatial pattern reconstruction and enhance generalization to unseen states. 
Moreover, a systematic analysis of reconstruction errors and potential physical violations under varying decoder performance represents an important direction for future work.

While the framework demonstrates broad applicability, its performance depends on several modeling assumptions. 
In particular, an adequate low-dimensional latent representation is crucial. 
If the system requires higher-dimensional embeddings, the increasing number of interaction terms may affect robustness and interpretability. 
Similarly, very high measurement noise can challenge inference and dynamics reconstruction, although the correct model terms may still be identified. 
The approach also builds on the assumption that latent dynamics can be captured in a sparse form; when this is not the case, solutions may be less interpretable or less efficient. 
Finally, a misspecified prior can bias the inference process and yield inaccurate uncertainty estimates.

Although the framework demonstrated robustness to noise, simulation-generated datasets cannot fully reflect the uncertainties, sensor imperfections, and modeling discrepancies inherent in real-world experiments. 
Consequently, dealing with more realistic measurement artifacts (e.g., impulsive, drift, or correlated noise) and applying the method to real-life datasets, including sensor and video data \cite{mars2022bayesian, chen2022automated}, represent exciting future directions to validate its practical robustness -- particularly given its non-intrusive nature.
As a further direction of development, the framework’s flexibility offers strong potential for online learning, allowing dynamic model updates as new data emerge. 
This could be achieved by using previous posterior estimates as priors and fine-tuning the model to refine posterior distributions, potentially in combination with conformal prediction techniques \cite{fasel2025sparse, stankeviciute2021conformal} to provide theoretically grounded uncertainty intervals under distribution shift and non-stationarity.


\FloatBarrier

\section*{Acknowledgment}
PC is supported under the JRC STEAM STM-Politecnico di Milano agreement and by the PRIN 2022 Project “Numerical approximation of uncertainty quantification problems for PDEs by multi-fidelity methods (UQ-FLY)” (No. 202222PACR), funded by the European Union - NextGenerationEU.

JK and JF are funded by Deutsche Forschungsgemeinschaft (DFG, German Research Foundation) under Germany’s Excellence Strategy - EXC 2075 - 390740016. We acknowledge the support by the Stuttgart Center for Simulation Science (SimTech).

AM acknowledges the project “Dipartimento di Eccellenza” 2023-2027 funded by MUR, the project FAIR (Future Artificial Intelligence Research), funded by the NextGenerationEU program within the PNRR-PE-AI scheme (M4C2, Investment 1.3, Line on Artificial Intelligence) and the Project “Reduced Order Modeling and Deep Learning for the real- time approximation of PDEs (DREAM)” (Starting Grant No. FIS00003154), funded by the Italian Science Fund (FIS) - Ministero dell'Università e della Ricerca. 

AF acknowledges the PRIN 2022 Project “DIMIN- DIgital twins of nonlinear MIcrostructures with iNnovative model-order-reduction strategies” 
(No. 2022XATLT2) funded by the European Union - NextGenerationEU.

SLB and JNK acknowledge generous funding support from the National Science Foundation AI Institute in Dynamic Systems
(grant number 2112085).

\subsection*{Statements and Declarations}

\paragraph*{Author contributions}
PC: conceptualization, data curation, investigation, methodology, software, validation, visualization, writing—original draft, writing-review and editing. \\
JK: conceptualization, data curation, investigation, methodology, software, validation, visualization, writing—original draft, writing-review and editing. \\
JF: conceptualization, funding acquisition, project administration, supervision, writing-review and editing. \\
AM: conceptualization, funding acquisition, project administration, supervision, writing-review and editing. \\
AF: conceptualization, funding acquisition, project administration, supervision, writing-review and editing. \\
JNK: conceptualization, funding acquisition, project administration, supervision, writing-review and editing. \\
SLB: conceptualization, funding acquisition, project administration, supervision, writing-review and editing. 

\paragraph*{Code and Data Availability}
The source code and data for the VENI, VINDy, VICI Framework, examples, and simulation setups are available at \url{https://github.com/jkneifl/VENI-VINDy-VICI} with a persistent release in~\cite{JonasKneifl2025}. All code is implemented in Python using TensorFlow. 


\bibliographystyle{elsarticle-num} 
\bibliography{references.bib}

\newpage
\appendix
\section{VENI–VINDy–VICI: methodological and implementation details}

\subsection{VINDy Loss}
\label{sec: vindy loss}
In this section we derive VINDy objective function \eqref{eq: VINDy Optimization} in Sect.~\ref{sec: VINDy}. We assume that $\sindyCoeff$ and $\latentState$ are independent random variables. This reflects the fact that $\sindyCoeff$ are constant coefficients determining the contribution of each candidate feature in the dynamics, thus they are related to the underlying physical model we aim to discover and not to the specific (latent) observed data.

In VINDy, the posterior distribution $p(\sindyCoeff|\dot{\latentState})$ is approximated with $\varPosterior(\sindyCoeff)$ parameterized by trainable weights. 
For the aforementioned reason, the approximated posterior for $\sindyCoeff$ does not feature an explicit dependency on the conditioning variable $\dot{\latentState}$.
The weights parameterizing $q$ are optimized by minimizing the KL divergence between the approximated posterior and the true one:
\begin{align}
\begin{split}
    \KL(\varPosterior(\sindyCoeff) &\parallel 
    \prob(\sindyCoeff \mid \dot{\latentState}))=\expectation_{\sindyCoeff \sim \varPosterior(\sindyCoeff)}[\log\varPosterior(\sindyCoeff)]
        -\expectation_{\sindyCoeff \sim \varPosterior(\sindyCoeff)}[\log\prob(\sindyCoeff \mid \dot{\latentState})] \\
    &=\expectation_{\sindyCoeff \sim \varPosterior(\sindyCoeff)}[\log\varPosterior(\sindyCoeff)]
        -\expectation_{\sindyCoeff \sim \varPosterior(\sindyCoeff)}[\log\prob(\sindyCoeff \mid \dot{\latentState},\, \latentState)] \\
    &= \expectation_{\sindyCoeff \sim \varPosterior(\sindyCoeff)}[\log\varPosterior(\sindyCoeff)]
        -\expectation_{\sindyCoeff \sim \varPosterior(\sindyCoeff)}
        \left[\log
            \frac{
                {
                \prob(\sindyCoeff,\, \latentState)%
                }    \prob(\dot{\latentState}  \mid \sindyCoeff,\, \latentState) 
            }
            {
                \prob(\dot{\latentState} \mid {\latentState})
                    \prob({\latentState})
            }
        \right]
     \\
     &= \expectation_{\sindyCoeff \sim \varPosterior(\sindyCoeff)}[\log\varPosterior(\sindyCoeff)]
        -\expectation_{\sindyCoeff \sim \varPosterior(\sindyCoeff)}
        \left[\log
            \frac
            {
                \prob(\sindyCoeff)\prob(\latentState)   \prob(\dot{\latentState}  \mid \sindyCoeff,\, \latentState) 
            }
            {
                \prob(\dot{\latentState} \mid {\latentState})
                    \prob({\latentState})
            }
        \right]
     \\
     &= \expectation_{\sindyCoeff \sim \varPosterior(\sindyCoeff)}[\log\varPosterior(\sindyCoeff)]
        -\expectation_{\sindyCoeff \sim \varPosterior(\sindyCoeff)}\left[\log\prob(\sindyCoeff)\right]  
        -\expectation_{\sindyCoeff \sim \varPosterior(\sindyCoeff)}\left[\log(\prob(\dot{\latentState}  \mid \sindyCoeff,\, \latentState))
        \right]
        + \log \prob(\dot{\latentState}\mid{\latentState})
     \\
     &= \KL(\varPosterior(\sindyCoeff) \parallel \prob(\sindyCoeff))
        -\expectation_{\sindyCoeff \sim \varPosterior(\sindyCoeff)}\left[\log(\prob(\dot{\latentState}  \mid \sindyCoeff,\, \latentState) )
        \right]
        + \log \prob(\dot{\latentState}\mid{\latentState}).
\end{split}
\label{eq:KL vindy}
\end{align}
The resulting expression has been obtained by applying KL divergence definition, Bayes' theorem on the second term and exploiting the independence between $\sindyCoeff$ and $\latentState$.\\
Analogously, the KL divergence between $p(\latentState|\dot{\latentState})$ and the approximated posterior $\varPosterior(\latentState)$ can be written as
\begin{align}
\begin{split}
    \KL(\varPosterior(\latentState) \parallel \prob(\latentState \mid \dot{\latentState}))
    &=  \expectation_{\latentState\sim \varPosterior(\latentState)}[\log\varPosterior(\latentState)]
        -\expectation_{\latentState\sim \varPosterior(\latentState)}[\log\prob(\latentState \mid \dot{\latentState})]\\ 
    &= \expectation_{\latentState\sim \varPosterior(\latentState)}[\log\varPosterior(\latentState)] 
        - \expectation_{\latentState\sim \varPosterior(\latentState)}[\log\prob(\latentState)]
        - \expectation_{\latentState\sim \varPosterior(\latentState)}[\log\prob(\dot{\latentState} \mid \latentState)] 
        + \log\prob(\dot{\latentState})\\
    &= \KL(\varPosterior(\latentState) \parallel \prob(\latentState))
        - \expectation_{\latentState\sim \varPosterior(\latentState)}[\log\prob(\dot{\latentState} \mid \latentState)] 
        + \log\prob(\dot{\latentState}).
 \end{split}
\label{eq:KL vindy 2}
\end{align}

\noindent Summing \eqref{eq:KL vindy} and \eqref{eq:KL vindy 2} up leads to
\begin{align}
\begin{split}
    \KL(\varPosterior(\sindyCoeff) \parallel\prob(\sindyCoeff \mid \dot{\latentState}))
    +&\KL(\varPosterior(\latentState) \parallel\prob(\latentState \mid \dot{\latentState}))\\
    =&  
    \,\KL(\varPosterior(\sindyCoeff) \parallel \prob(\sindyCoeff))
        -\expectation_{\sindyCoeff \sim \varPosterior(\sindyCoeff)}
        \left[
            \log(\prob(\dot{\latentState}  \mid \sindyCoeff,\, \latentState) )
        \right]
        + \log \prob(\dot{\latentState}\mid{\latentState}) \,+       \\
        &\,\KL(\varPosterior(\latentState) \parallel \prob(\latentState))
        - \expectation_{\latentState\sim \varPosterior(\latentState)}[\log\prob(\dot{\latentState} \mid \latentState)] 
        + \log\prob(\dot{\latentState}) \\
    =&  
    \,\KL(\varPosterior(\sindyCoeff) \parallel \prob(\sindyCoeff))
        + \KL(\varPosterior(\latentState) \parallel \prob(\latentState))
        -\expectation_{\sindyCoeff \sim \varPosterior(\sindyCoeff)}
        \left[
            \log(\prob(\dot{\latentState}  \mid \sindyCoeff,\, \latentState))
        \right]
        + \log\prob(\dot{\latentState}).
 \end{split}
\label{eq:KL sum}
\end{align}
The last equality follows from the standard assumption of  stochastic gradient descent-based methods: at each optimization step, we take a sample of $\latentState$ and treat $\log \prob(\dot{\latentState}|\latentState)$ as an approximation of $\expectation_{\latentState\sim\varPosterior(\latentState)}[\log \prob(\dot{\latentState}|\latentState)]$, thus assuming $\log \prob(\dot{\latentState}|\latentState)-\expectation_{\latentState\sim\varPosterior(\latentState)}[\log \prob(\dot{\latentState}|\latentState)]\approx 0 $.\\
\noindent Finally, equation \eqref{eq:KL sum} can be rewritten as
\begin{align*}
 \begin{split}
    \log \prob(\dot{\latentState}) 
    &- \KL(\varPosterior(\sindyCoeff) \parallel \prob(\sindyCoeff|\dot{\latentState})) 
    - \KL(\varPosterior(\latentState) \parallel \prob(\latentState|\dot{\latentState})){\approx}
    \\
    &{\approx} \expectation_{\sindyCoeff \sim \varPosterior(\sindyCoeff)}
        \left[
            \log(\prob(\dot{\latentState}  \mid \sindyCoeff, \latentState))
        \right] - \KL(\varPosterior(\sindyCoeff) \parallel \prob(\sindyCoeff))
        - \KL(\varPosterior(\latentState) \parallel \prob(\latentState)),
 \end{split}
\end{align*}
that is exactly the VINDy objective function in \eqref{eq: VINDy Optimization}.


\subsection{Prior Distributions}
\label{app: prior}

The priors~$\prob(\sindyCoeff)$ and $\prior$ in the context of VENI and VINDy can be used to infuse knowledge about the distributions of the coefficients of the dynamics and latent states. If no prior knowledge about the distribution's shape and its parameters is present, Gaussian priors are a common choice in variational inference, especially in VAEs. Moreover, using latent Gaussian distributions for the latent variables allows modeling arbitrarily complex output distributions if the decoder is sufficiently complex \cite{devroye1986sample}. 
A valid alternative, especially when modeling the dynamics' coefficients is the use of Laplacian priors. In contrast to Gaussian priors, Laplacian priors can act as sparsity-promoting regularization terms within a regression task \cite{Hirsh2022sparsifying}. Consequently, they are more prone to provide a parsimonious dynamical model, which is in general highly desirable {\cite{kutz2022parsimony}.
We furthermore want to highlight that adjusting the priors are an elegant way of incorporating any existing pre-knowledge about occurring function terms. 
Other possible choices for the priors may include spike and slab, regularized horseshoe, etc.
In this work, we focus on Gaussian and Laplacian priors, since they have the additional advantage that the KL divergence between two distribution of the same family can be expressed in closed form, as presented in the following.

\paragraph*{Gaussian}
We denote by
\begin{align}
\NormalDist(\GaussianMean, \GaussianVar)
\label{eq:1}
\end{align}
the Gaussian distribution with mean~$\GaussianMean$ and variance~$\GaussianVar$.
For two univariate Gaussian distributions, the KL divergence can be calculated in closed form as
\begin{align}
\KL(\NormalDist(\GaussianMean_1, \GaussianVar_1) \parallel \NormalDist(\GaussianMean_2, \GaussianVar_2))
&= \log\left(\frac{\GaussianVar_2}{\GaussianVar_1}\right)
+ \frac{{\GaussianVar_1}^2 + (\GaussianMean_2-\GaussianMean_1)^2}
{2{\GaussianVar_2}^2} - \tfrac{1}{2}.
\end{align}


\paragraph*{Laplace}
We denote by
\begin{align}
\LaplacianDist(\locationParam, \scaleParam)
\end{align}
the Laplace distribution with location parameter~$\locationParam$ and scale parameter~$\scaleParam$.
For gradient-based optimization the reparameterization trick for a Laplace distribution looks similar as the one for Gaussian distributions.
\begin{align}
    \latentState = \locationParam + \vaeNoise\scaleParam, \quad \vaeNoise\sim\LaplacianDist(0, 1).
\end{align}
Given two Laplace distributions, the KL~divergence can be calculated in closed form as
\begin{align}
    \KL(\LaplacianDist(\locationParam_1, \scaleParam_1) \parallel \LaplacianDist(\locationParam_2, \scaleParam_2))= \log(\frac{\scaleParam_2}{\scaleParam_1}) + \frac{\scaleParam_1\exp{-\frac{\vert \locationParam_1-\locationParam_2\vert }{\scaleParam_1}} + \vert \locationParam_1-\locationParam_2\vert}{\scaleParam_2} - 1,
\end{align}
see~\cite{Meyer2021CVPR}.

\subsection{Approach to Solving the Joint Optimization of VENI and VINDy}
\label{app: solve_veni_vindy}
Thanks to the assumptions and choices on the family of distributions, the general, probabilistic objective \eqref{eq: loss} can be translated into the practical, data-driven optimization problem \eqref{eq: loss_data}, where the expectation is approximated by the sample mean over the training set. 
The \textit{reconstruction} loss in \eqref{eq: loss} is proportional to the squared Euclidean distance between the data $\states^{(i)}$ and the decoder mean $\hat{\states}^{(i)}=\vaeMean(\latentState^{(i)};\weights_{\decoder})$, since we assumed isotropic Gaussian distribution for the decoder (see \eqref{eq: decoder_dist}) \cite{yu2020tutorial}. 
For the exact same reason, as the posterior distribution of $\dot{\latentState}$ has the same characteristics (see \eqref{eq: dz_dist}), the \textit{latent dynamics} loss can be rewritten here as proportional to ${\norm{\dot{\latentState}^{(i)} - \sindyCoeff^{(i)}\sindyLib(\latentState^{(i)},\param)}}_2^2$. Moreover, as suggested in \cite{champion2019data}, an additional \textit{full-dynamics} term is included as regularization to account for consistency between the real time derivatives of the data, $\dot{\states}^{(i)}$, and the mean value of their reconstruction from the approximated latent derivatives, that is $\dot{\hat{\states}}^{(i)}\coloneqq\nabla_{\latentState}\vaeMean^{(i)}\sindyCoeff^{(i)}\sindyLib(\latentState^{(i)},\param)$. Finally, given the assumption that the families of distributions for the posteriors $q(\sindyCoeff|\dot{\latentState}^{(i)})$ and $q(\latentState|\states^{(i)})$ (that is $\encoder(\states^{(i)})$) are the same of those of the priors (in the cases of Gaussian and Laplacian distributions), the KL divergences in \eqref{eq: loss_data} can be expressed in closed-form and computed directly from the statistical moments of the posteriors (see \ref{app: prior}). By minimizing the divergences, the posteriors are pushed closer to the selected priors which helps minimizing the optimization goal formulated in \eqref{eq: objective2} and \eqref{eq: VINDy Optimization}.  

As a general guideline for selecting the hyperparameters~$\{\lambda_i\}_{i=1}^{5}\in \R^+$ in \eqref{eq: loss_data}, the dominant terms are those related to the \textit{reconstruction} and identification of the \textit{latent dynamics}, \i.e., $\lambda_1$ and $\lambda_3$. Thus, they should be kept orders of magnitude larger than $\lambda_2$, $\lambda_4$, and $\lambda_5$, which weight the remaining regularization terms. 
In some cases, the method may converge to degenerate solutions, e.g., trivial dynamics such as~$\dot{\latentState}=\mathbf{0}$. 
Here, the \textit{latent dynamics} loss can appear deceptively low -- not because the dynamics are correctly identified, but simply due to latent variables with vanishing magnitudes.
To address this, $\lambda_5$ can be increased so that the \textit{full dynamics} term ensures the identified dynamics are respected on the original variables, thus mitigating this scaling issue. 
Please note that computing the full dynamics loss can be computationally demanding, as it requires evaluating the decoder Jacobian -- particularly in large-scale systems. 
To alleviate this, a linear dimensionality-reduction step (via PCA) prior to the nonlinear decoder can be implemented into the framework. 
This substantially lowers the dimensionality of the processed state variables and ensures that backpropagation through the latent states remains tractable.

Regarding the KL-losses, a prioritized weighting of these terms might be used, as done for $\beta-$VAEs~\cite{higgins2016beta}, to encourage more efficient latent encoding and disentanglement of the latent coordinates.
In general, the precise hyperparameter values need to be adjusted to the specific dataset at hand, as the choice of hyperparameters is problem-dependent and their regulation could improve the training convergence. 
In addition, it would also be possible to use constrained learning~\cite{Chamon2020} so that, for example, dynamics are only learned in an area where satisfactory reconstruction is achieved.

\subsection{Network Implementation}
\label{sec: network_implementation}
In this secton, we report the implementation details of the VENI–VINDy–VICI framework, including the network architectures, training configurations, and dataset specifications used for the numerical experiments. 
Table~\ref{tab:network hyperparameter} summarizes the main hyperparameters, latent priors, optimization settings, and loss weighting factors for each case study.
Table~\ref{tab:data hyperparameter} provides an overview of the corresponding dataset configurations, including training and testing set sizes, time horizons, and signal-to-noise ratios for the measured states and their derivatives. These details ensure the reproducibility of the results and clarify the computational setup used in the presented experiments.

\begin{table}[htb]
	\setlength{\tabcolsep}{3pt}
	\centering
	\caption{Summary of network architecture hyperparameters}
	\label{tab:network hyperparameter}
	\npdecimalsign{.}
	\nprounddigits{2}
	\begin{tabular}[c]{l l l l}
		\toprule
        &\textbf{R\"ossler}& \textbf{Reaction-diffusion} & \textbf{MEMS resonator} \\
		\midrule
		\textbf{latent dim.} $\latentStateDim$ &3 & 2 & 1\\
		\textbf{PCA dim.} & - & 32 & 3\\
        \midrule
        \textbf{encoder layer sizes} & - & [32, 16, 8] & [32, 32, 32]\\
        \textbf{decoder layer sizes} & - & [8, 16, 32] & [32, 32, 32]\\
        \textbf{latent priors for} $\latentState$& - & $\NormalDist (\bm{0},\bm{I})$ & $\NormalDist (\bm{0},\bm{I})$\\
        \textbf{activation function} & - & elu & elu \\
        \midrule
        \textbf{coefficient priors for} $\sindyCoeff$& $\LaplacianDist(0, 1)$ &  $\LaplacianDist(0, 1)$ & $\LaplacianDist(0, 1)$\\
        \textbf{PDF threshold} $pdf_{\sindyCoeff}$ & 5 & 5 & 1.9 \\
        \midrule
        \textbf{loss factors}  \\
        $\quad$ reconstruction $\lambda_{1}$ & - & 1e-2 & {1e-3} \\
        $\quad$ posterior~$\latentState$ $ \lambda_{2}$ & - & {2e-5} & {1e-8} \\
        $\quad$ latent dynamics $\lambda_{3}$ & 1 & 4 & {1} \\
        $\quad$ posterior~$\sindyCoeff$ $ \lambda_{4}$ & {1e-3} & {1e-4} & {1e-8} \\
        $\quad$ full dynamics $ \lambda_{5}$ & - & {1e-2} & {1e-5} \\
        \textbf{epochs} & 500 & 2000 & 2500 \\
        \textbf{batch size} & 256 & 256 & 256 \\
        \textbf{optimizer} & Adam & Adam & Adam \\
        \textbf{learning rate} & {1e-3} & {1e-3} & {2e-3} \\
        \midrule
        \textbf{time per epoch}
            & 0.68\,s 
            & 0.57\,s
            & 4.32\,s \\
        \textbf{inference time}\\
        $\quad$ time integration + decoding 
            & 0.1994\,s  
            & 0.79\,s 
            & 2.08\,s  \\
        $\quad$ {per simulated time unit}
            & 0.0083
            & 0.0020
            & 0.0012  \\
        \bottomrule
	\end{tabular} 
\end{table}

\begin{table}[htb]
	\setlength{\tabcolsep}{3pt}
	\centering
	\caption{Summary of dataset configurations.}
	\label{tab:data hyperparameter}
	\npdecimalsign{.}
	\nprounddigits{2}
	\begin{tabular}[c]{l l l l}
		\toprule
        \textbf{data shape} \\
        $\quad$ training $[N_\beta^{\text{train}} \times N_t^{\text{train}} \times \stateDim]$
            & $[30 \times 2000 \times 3]$ 
            & $[16 \times 400 \times 2500]$ 
            & $[28 \times 14001 \times 7821]$ \\
        $\quad$ test $[N_\beta^{\text{test}} \times N_t^{\text{test}} \times \stateDim]$ 
            & $[4 \times 2000 \times 3]$ 
            & $[4 \times 800 \times 2500]$ 
            & $[28 \times 22499 \times 7821]$ \\
        \midrule
        \textbf{time window}\\
        $\quad$ training
            & [0, 24]\,s 
            & [0, 20]\,s
            & [0, 1091]\,$\mu\text{s}$ \\
        $\quad$ inference
            & [0, 24]\,s  
            & [0, 40]\,s
            & [0, 1755]\,$\mu\text{s}$  \\
        \midrule
        \textbf{signal-to-noise-ratio}\\
            $\quad$ state $\states$ & $\infty$/26/20\,dB & 14\,dB & 38\,dB \\
            $\quad$ velocity $\dot{\states}$ & $\infty$/-5/-11\,dB & 0.4\,dB & 13\,dB \\
            $\quad$ acceleration $\ddot{\states}$ & - & - & -12\,dB \\
		\bottomrule
	\end{tabular} 
\end{table}

\section{Benchmark comparison}
\label{sec: comparison}
\setcounter{table}{0} 

To clarify how our proposed method compares to existing alternatives, we benchmark the VENI, VINDy, VICI (V-V-V) framework against two representative approaches: the Fourier Neural Operator (FNO) \cite{li2020fourier} and a Gaussian Process (GP)-based reduced-order model. 
Neural operators such as FNOs have achieved state-of-the-art performance in solving parametric PDE problems, by efficiently learning mappings between infinite-dimensional function spaces.
In contrast, the GP-based ROM configuration was included as a reference for probabilistic modeling, as GPs represent a well-established framework for uncertainty quantification. 
In particular, we consider GP regression within an autoencoder framework (AE-GP), enabling probabilistic modeling of the latent dynamics while retaining a reduced-order representation of the system.

Table~\ref{tab:comparison_methods} summarizes the key characteristics of the proposed framework and the two benchmark approaches across several relevant criteria, including uncertainty quantification, computational efficiency, scalability, accuracy, extrapolation performance, hyper-resolution capabilities, and physical interpretability.

\begin{table}[b!]
\centering
\caption{Comparison of the proposed V-V-V framework and benchmark methods (FNO and AE+GP) across key properties relevant for reduced-order modeling of dynamical systems. 
A checkmark (\checkmark) indicates that the method possesses the property, a cross (\text{\sffamily X}) indicates the property is absent, and dash ($-$) indicates partial fulfillment. For temporal hyper-resolution, a mixed entry (X/\checkmark) denotes that the property depends on the specific implementation (see Paragraph \textit{Zero-shot hyper-resolution (space/time)}).}
\label{tab:comparison_methods}
\renewcommand{\arraystretch}{1.2}
\begin{tabular}{l|c|c|c}
\hline
\textbf{Property} & \textbf{V-V-V} & \textbf{FNO} & \textbf{AE+GP} \\
\hline
\textbf{Uncertainty quantification} & \checkmark & \text{\sffamily X} & \checkmark \\
\textbf{Offline training time} & \checkmark & $-$ & \text{\sffamily X} \\
\textbf{Online inference time} & \checkmark & \checkmark & \text{\sffamily X} \\
\textbf{Scalability} & \checkmark & \checkmark & \text{\sffamily X} \\
\textbf{In-domain accuracy} & \checkmark & \checkmark & \text{\sffamily X} \\
\textbf{Extrapolation accuracy} & \checkmark & $-$ & \text{\sffamily X} \\
\textbf{Spatial zero-shot hyper-resolution} & \text{\sffamily X} & \checkmark & \text{\sffamily X} \\
\textbf{Temporal zero-shot hyper-resolution} & \checkmark &  \text{\sffamily X}/\checkmark &  \text{\sffamily X}/\checkmark\\
\textbf{Physical interpretability} & \checkmark & \text{\sffamily X} & \text{\sffamily X} \\
\hline
\end{tabular}
\end{table}

\paragraph{Uncertainty quantification}  
Both the proposed framework and the AE+GP benchmark provide explicit UQ mechanisms, whereas FNOs lack native UQ capabilities. 
In the proposed framework, uncertainty is modeled at multiple levels: VENI quantifies uncertainty in the system states, VINDy in the latent dynamics, and VICI in the inferred model predictions. Similarly, GP-based approaches can provide UQ on both states and dynamics when formulated in a state-space setting (see, e.g., \cite{eleftheriadis2017identification}), offering a probabilistic reference for uncertainty modeling in dynamical systems.

\paragraph{Computational efficiency}  
To quantify computational efficiency, we tested V-V-V, FNO, and AE+GP on the MEMS resonator example described in Sect.~\ref{sec: results_MEMS}.  

For the FNO implementation, we use a standard architecture performing Fourier transforms in space and autoregressive propagation in time \cite{li2020fourier}, trained to map the solution at the previous $n_\text{in}=5$ time steps to the subsequent $n_\text{out}=3$ steps. Input and output channels correspond to the concatenation of state and forcing variables over the bundled time steps, and the model uses $n_\text{modes}=14$ Fourier modes and $n_\text{hidden}=10$ hidden channels per layer.\\  
For the AE-GP implementation, high-dimensional states are first projected onto a reduced latent space using an autoencoder with two hidden layers of 16 neurons. A variational Gaussian process with inducing points, implemented via \texttt{GPyTorch} \cite{gardner2018gpytorch}, is trained in a state-space manner to map the current latent state and system parameters to the next-step latent state. Predictions are propagated autoregressively and decoded back to the original space.

\begin{table}[b!]
\centering
\renewcommand{\arraystretch}{1.2}
\caption{Comparison of computational costs for the MEMS resonator example (Sect.~\ref{sec: results_MEMS}). V-V-V and FNO have similar inference times, but FNOs require substantially more training time per epoch, while AE+GP is considerably slower for both training and inference. All times were measured on a workstation with an AMD Ryzen 9 5950X 16-core CPU.}
\label{tab:beam_quantitative}
\begin{tabular}{l|c|c|c}
\hline
\textbf{Metric} & \textbf{V-V-V} & \textbf{FNO} & \textbf{AE+GP} \\
\hline
\textbf{Training time per epoch} & 4.3\,s & 46\,s & 85\,s \\
\textbf{Inference time per trajectory} & 2.08\,s & 2.58\,s & 22\,s \\
\hline
\end{tabular}
\end{table}

V-V-V and FNO exhibit comparable inference times, whereas AE+GP models are significantly more expensive, requiring roughly an order of magnitude more time per trajectory (see Table~\ref{tab:beam_quantitative}). 
FNOs are more costly to train per epoch than V-V-V, primarily due to the FFT-based spectral convolutions, however both V-V-V and FNO can handle high-dimensional inputs and long trajectories by could benefit from GPU-acceleration and scalable mini-batch optimization. In contrast, AE+GP approaches, while reducing dimensionality via the autoencoder, remain constrained by the computational cost of GP regression, limiting both training efficiency and scalability.

\paragraph{Scalability}
In contrast to many autoregressive schemes such as discrete state-space Gaussian Processes (GPs), our approach is not tied to a fixed step size for the latent state evolution. 
Instead, adaptive numerical solvers determine the integration step, which allows us to coarsen training data, for instance by down-sampling in phases of low dynamics, without compromising consistency at inference. 
This flexibility reduces the computational burden for large datasets, while training remains straightforward and scalable through conventional gradient-based optimization with mini-batching.

Moreover, the memory footprint of our V-V-V scales only linearly with the number of training samples, in contrast to Gaussian Process models, whose memory demand grows quadratically and whose inference cost scales cubically with the dataset size. 

\paragraph{Accuracy and extrapolation}
The proposed V-V-V framework attains a mean relative error of approximately 6.4\% (averaged over space and time) on the testing trajectories, i.e., for unseen parameters and extrapolated time instances, while for FNOs the error increases to 15.1\%. The higher error observed in FNOs arises from their limited extrapolation capability, as they rely on autoregressive predictions that accumulate errors over long horizons. In contrast, V-V-V maintains predictive accuracy beyond the training window thanks to its explicit physical latent dynamical model. AE+GP models fail to provide stable autoregressive rollouts for testing and even training parameters, diverging quickly when propagated in time, even within the training time window.


\paragraph{Zero-shot hyper-resolution (space/time)}
Zero-shot hyper-resolution refers to the ability to predict solutions at finer spatial or temporal resolutions than those used during training, without retraining.
The proposed framework enables temporal hyper-resolution by explicitly identifying a continuous latent dynamical model.
In contrast, FNO and AE+GP with standard autoregressive implementations are tied to the training time-step and cannot meaningfully refine temporal resolution.
FNOs that perform Fourier transforms in both space and time could, in principle, allow temporal hyper-resolution, but this ties the forecasting horizon to the training time window, limiting extrapolation.
GP-based models treating time as an explicit input could also allow finer resolution, but the cubic complexity of GP inference makes small time-step predictions infeasible for long trajectories.

Spatial hyper-resolution is naturally supported only by FNOs, while both our method and GP are tied to the training spatial resolution.
This limitation highlights an exciting opportunity for our framework: its modular design could allow resolution-invariant architectures by replacing the standard AE component \cite{bunker2025autoencoders}.

\paragraph{Physical interpretability}
The proposed V-V-V framework provides an explicit latent dynamical system that recovers the underlying governing equations, offering a physically and dynamically interpretable representation of the observed system. In contrast, both FNO and AE+GP operate as black-box input–output mappings. Moreover, autoregressive formulations that use multiple time steps as input for rollout may be inconsistent with the initial value problem of PDEs, where trajectories are determined solely by the initial condition.


Overall, the comparison highlights how the V-V-V framework uniquely combines uncertainty quantification, scalability, and accuracy, while supporting extrapolation and providing physical interpretability. Quantitative results further confirm its robust long-horizon and extrapolation performance, emphasizing its advantages over alternative state-of-the-art methods.

\end{document}

%% file: variables.tex
\usepackage{amssymb,amsmath,amsthm,amsbsy, mathtools, bm}

\newcommand{\Rdim}{\ensuremath\mathbb{R}}
\newcommand{\transpose}{\intercal}

\newcommand{\states}{\bm{x}}
\newcommand{\dataset}{\vb{X}}
\newcommand{\stateDim}{N}
\newcommand{\latentState}{\bm{z}}
\newcommand{\latentStateDim}{n}

\newcommand{\param}{\bm{\beta}}
\newcommand{\R}{\mathbb{R}}

\newcommand{\autoencdoer}{\bm{\phi}}
\newcommand{\encoder}{\autoencdoer_{\text{e}}}
\newcommand{\decoder}{\autoencdoer_{\text{d}}}
\newcommand{\weights}{\bm{W}}
\newcommand{\weightsentry}{W}

\newcommand{\prob}{p}
\newcommand{\posterior}{\prob(\latentState|\states)}
\newcommand{\marginal}{\prob(\states)}
\newcommand{\prior}{\prob(\latentState)}
\newcommand{\varPosterior}{q}
\newcommand{\varPosteriorFamily}{\mathcal{Q}}
\newcommand{\KL}{\text{KL}}

\newcommand{\expectation}{\mathbb{E}}
\newcommand{\vaeMean}{\bm{\hat{\states}}}
\newcommand{\vaeVar}{\bm{\delta}}

\newcommand{\vaeNoise}{\bm{\zeta}}

\newcommand{\NormalDist}{\mathcal{N}}
\newcommand{\LaplacianDist}{\mathcal{L}}
\newcommand{\standardDeviation}{\sigma}
\newcommand{\GaussianMean}{{\mu}^{\NormalDist}}
\newcommand{\GaussianVar}{{\delta}^{\NormalDist}}
\newcommand{\locationParam}{{\mu}^{\LaplacianDist}}
\newcommand{\scaleParam}{{\delta}^{\LaplacianDist}}
\newcommand{\pdfThreshold}{\tau}

\newcommand{\sindyCoeff}{\bm{\Xi}}
\newcommand{\sindyCoeffentry}{\Xi}
\newcommand{\sindyLib}{\bm{\Theta}}
\newcommand{\sindyParam}{\bm{\beta}}

\newcommand{\noise}{\bm{\varepsilon}}
\newcommand{\modelParam}{\alpha}
\newcommand{\modelParams}{\bm{\alpha}}
\newcommand{\roeA}{z_1}
\newcommand{\roeB}{z_2}
\newcommand{\roeC}{z_3}
\newcommand{\roeCoeffA}{\modelParam_1}
\newcommand{\roeCoeffB}{\modelParam_2}
\newcommand{\roeCoeffC}{\modelParam_3}
\newcommand{\modelNoiseFactor}{\lambda_{\modelParam}}
\newcommand{\measNoiseFactor}{\lambda_{\noise}}

\newcommand{\micr}{\,$\mu$m\,}

\newcommand{\bfm}[1]{\mathbf{#1}}

\newcommand{\bfzero}{\bfm{0}}

\newcommand{\bfB}{\bfm{B}}

\newcommand{\bfu}{\bfm{u}}

\newcommand{\bfP}{\bfm{P}}

\newcommand{\bfN}{\bfm{N}}

%% file: Figures/figure_config.tex
\usepackage{bm}
\usepackage{amsmath}
\usepackage{xcolor}
\usepackage{pgfplots}
\usepgfplotslibrary{fillbetween}
\usetikzlibrary{calc, 3d, positioning, fit, shapes.geometric, math, arrows.meta, fadings, through, quotes}
\pgfplotsset{compat=1.18} 

\definecolor{phColor}{RGB}{255, 131, 131}

\definecolor{stateColor}{RGB}{0,104,163}
\definecolor{aeColor}{RGB}{60, 116, 200}
\definecolor{latentColor}{RGB}{252, 114, 255}
\definecolor{VariationalColor}{RGB}{164,216,91}
\colorlet{weightColor}{phColor}

\definecolor{testColor}{rgb}{0.1, 0.1, 0.1}
\definecolor{trainColor01}{RGB}{55, 139, 205}
\definecolor{trainColor0}{RGB}{95, 189, 255}
\definecolor{trainColor1}{RGB}{255, 175, 128}
\definecolor{trainColor11}{RGB}{220, 123, 83}
\definecolor{trainColor2}{RGB}{234, 95, 130}
\definecolor{trainColor21}{RGB}{164, 51, 86}
\definecolor{trainColor3}{RGB}{106, 44, 112}
\definecolor{trainColor31}{RGB}{106, 44, 112}

\tikzset{
        cell/.style={
                rectangle, 
                rounded corners=.25mm, 
                minimum height =0.7cm, 
                minimum width=0.7cm, 
                draw=gray,
                anchor=center,
        },
        trapez/.style={
                trapezium, 
                rounded corners=0.15mm,
                minimum height=12mm, 
                trapezium stretches=true,
                trapezium angle=60, 
                inner sep=0mm,
                draw=gray,
                shape border rotate=270, 
        },
        mylabel/.style={%
                align=center,
                text=darkgray,
        },      
        Arrow/.style={
                -stealth,
                rounded corners=.1cm,
                thick,
                draw=gray,
                shorten >= 3pt,
                shorten <= 3pt
        },
        neuron/.style={circle, draw, minimum size=4ex, thick},
        operator/.style={circle, minimum width=7mm, fill=lightgray!40!white, draw=lightgray},
        equation/.style={cell},
}

\pgfdeclarelayer{bg0}    
\pgfdeclarelayer{bg1}    
\pgfsetlayers{bg1, bg0, main}

\newcommand\denselyConnectNodes[2]{

  \begin{pgfonlayer}{bg0}
  \foreach \n [count=\lyrIdx, remember=\lyrIdx as \previdx, remember=\n as \prevn] in #2 {
    \foreach \y in {1,...,\n} {
      \ifnum \lyrIdx > 1
        \foreach \x in {1,...,\prevn}
          \draw[-, draw=gray, 
          shorten >= 10pt,
          shorten <= 10pt, 
          thick] (#1-\previdx-\x.center) -- (#1-\lyrIdx-\y.center);
      \fi
    }
  }
  \end{pgfonlayer}
}


%% file: Figures/vici/vici.tex



\newcommand{\stateOpacity}{1}
\newcommand{\stateOpacityRec}{1}
\newcommand{\pcaOpacity}{1}
\newcommand{\pcaOpacityRec}{1}
\newcommand{\intermediateOpacity}{1}
\newcommand{\intermediateOpacityRec}{1}
\newcommand{\aeOpacity}{1}
\newcommand{\aeOpacityRec}{1}
\newcommand{\latentOpacity}{1}
\newcommand{\surrogateOpacity}{1}

\newcount\hidereconstruction
\newcount\hidereduction
\newcount\hideph
\newcount\hideae
\newcount\aeOnly
\newcount\hidelinear
\hidereconstruction=0
\hidereduction=0
\hidelinear=0
\hideae=0
\aeOnly=0
\hideph=0
\ifnum\hidereconstruction=1
    \renewcommand{\stateOpacityRec}{0.3}
    \renewcommand{\pcaOpacityRec}{0.3}
    \renewcommand{\intermediateOpacityRec}{0.3}
    \renewcommand{\aeOpacityRec}{0.3}
\fi
\ifnum\hidelinear=1
    \renewcommand{\stateOpacity}{0.3}
    \renewcommand{\pcaOpacity}{0.3}
    \renewcommand{\stateOpacityRec}{0.3}
    \renewcommand{\pcaOpacityRec}{0.3}
    \renewcommand{\intermediateOpacityRec}{0.3}
\fi
\ifnum\hideae=1
    \renewcommand{\aeOpacityRec}{0.3}
    \renewcommand{\aeOpacity}{0.3}
    \renewcommand{\latentOpacity}{0.3}
\fi
\ifnum\hidereduction=1
    \renewcommand{\stateOpacity}{0.3}
    \renewcommand{\pcaOpacity}{0.3}
    \renewcommand{\intermediateOpacity}{0.3}
    \renewcommand{\aeOpacity}{0.3}
\fi
\ifnum\hideph=1
    \renewcommand{\surrogateOpacity}{0.3}
\fi
\ifnum\aeOnly=1
    \renewcommand{\stateOpacity}{0.00}
    \renewcommand{\pcaOpacity}{0.0}
    \renewcommand{\stateOpacityRec}{0.0}
    \renewcommand{\pcaOpacityRec}{0.0}
    \renewcommand{\intermediateState}{\states}
\fi

\begin{tikzpicture}[
    shorten >=1pt, shorten <=1pt,
    pca_opacity/.style={opacity=\pcaOpacity,},
    pca_opacity2/.style={opacity=\pcaOpacityRec,},
    ae_opacity/.style={opacity=\aeOpacity,},
    ae_opacity2/.style={opacity=\aeOpacityRec,},
    state_opacity/.style={opacity=\stateOpacity,},
    state_opacity2/.style={opacity=\stateOpacityRec,},
    intermediatestate_opacity/.style={opacity=\intermediateOpacity,},
    intermediatestate_opacity2/.style={opacity=\intermediateOpacityRec,},
    latent_opacity/.style={opacity=\latentOpacity,},
    surrogate_opacity/.style={opacity=\surrogateOpacity,},
    x={(1cm,0cm)},y={(0cm,1cm)},z={(0.8cm,0.2cm)}
  ]

    \newcommand{\sepSpace}{.75mm}

    
    \node[inner sep=0pt,
        minimum width=16mm,
        minimum height=16mm,
            ] (state)
        {\includegraphics[height=23mm]{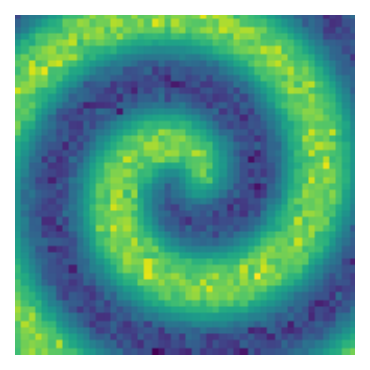}};

    \node [trapez,
            pca_opacity,
            minimum width=23mm, 
            minimum height=15mm,
            trapezium angle=50, 
            draw=aeColor!80!black,  
            fill=aeColor!20!white,
            anchor=west,
            label={
            [label distance=-0.75mm, pca_opacity, mylabel, rotate=-28]
            above:}
        ] at ($(state.east) + (3*\sepSpace, 0)$) (encoder) {$\encoder$};


    \node [cell, 
    label={[label distance=9.5mm, latent_opacity, mylabel]:},
        draw=VariationalColor!80!black,
        pca_opacity,
        minimum height=7mm, 
        minimum width=20mm, 
        latent_opacity,
        anchor=west,
        right=2.5*\sepSpace of encoder.east,
        fill=VariationalColor!20!white] (latent dist) {$\qquad \sim \latentState_0^{(i)} $};
        \begin{scope}[x=6.5mm,y=6mm, shift={($(latent dist.center)-(0.7,0)$)}, scale=0.15]
          \draw[gray!50,thick] (-3.5,-2.25) -> (3.5,-2.25);
          \draw[gray!50,thick] (0,-2.2cm) -> (0,2.cm);
          \draw[thick] (0,0) plot[domain=-3.5:3.5] (\x, {1/(0.1*2.5) * exp(-(\x*\x/2))) - 2.25 });
        \end{scope}

    \node [cell, 
            draw=VariationalColor!80!black,
            fill=VariationalColor!20!white,
        minimum height=7mm, 
        minimum width=20mm, 
            surrogate_opacity, 
            label={[label distance=0mm, surrogate_opacity, mylabel]below:}]
            at ($ (latent dist.center) + (0,0.8) $) (vindyDist) {$\qquad  \sim \sindyCoeff^{(i)} $};
            \begin{scope}[x=6.5mm,y=6mm, shift={($(vindyDist.center)-(0.7,0)$)}, scale=0.15]
                  \draw[gray!50,thick] (-3.5,-2.25) -> (3.5,-2.25);
                    \draw[gray!50,thick] (0,-2.2cm) -> (0,2.cm);
                  \draw[thick] (0,0) plot[domain=-3.5:3.5] (\x, {1/(2*0.15) * exp(-abs(\x)/0.6)) -2 });
            \end{scope}
    
    \node [cell,
            draw=latentColor!80!black,
            anchor=south,
            minimum width=7mm,
            minimum height=7mm,
            surrogate_opacity,
            fill=latentColor!20!white,
            label={[label distance=0mm, surrogate_opacity, mylabel]}] 
            at ($ (state.north) + (0,.1) $) (param) {$\sindyParam$};

                
    \node [cell, 
            draw=none,
            anchor=west,
            minimum height=9.5mm, 
            surrogate_opacity, 
            label={[label distance=0mm, surrogate_opacity, mylabel]below:}]
            at ($ (vindyDist.center) + (1.5, -0.4) $) (ode) {$    
                \begin{cases} 
                    \vspace{2mm}
                    \dot{\latentState} = \sindyCoeff^{(i)}\sindyLib(\latentState,\param)\\ 
                    \latentState(0) = \latentState^{(i)}_0,
                \end{cases}$};

                
    \node [cell, 
            draw=none,
            minimum height=9.5mm, 
            surrogate_opacity, 
            label={[label distance=0mm, surrogate_opacity, mylabel]below:}]
            at ($ (ode) + (-1, -2.75) $) (trajectory) {$\{\latentState^{(i)}(t): t\in\mathcal{T}\}_{i=1}^{m}$};
            
	    \node [ 
        draw=none,
        fill=none,
        minimum height=7mm, 
        minimum width=30mm, 
			] at ($ (trajectory.center) + (0, -0.85) $)
            (latent_trajectory) {};
            \begin{scope}[x=100mm,y=20mm, shift={($(latent_trajectory.west)$)}, scale=.05]
                \draw[thick, draw=gray] (0,0) plot[domain=0:6, samples = 100] ({1.05*\x-.2}, {-4});
                \draw[thick, draw=gray] (0,0) plot[domain=0:6, samples = 100] (0, {1.5*\x-5});
                \draw[thick, draw=latentColor] (0,0) plot[domain=0:6, samples = 100] (\x, {exp(-sin(\x*200)/0.6) -2 });
                \draw[thick, draw=latentColor!20!white, name path = A] (0,0) plot[domain=0:6, samples = 100] (\x, {exp(-sin(\x*200)/0.6) -2 + 0.3*\x});
                \draw[thick, draw=latentColor!20!white, name path = B] (0,0) plot[domain=0:6, samples = 100] (\x, {exp(-sin(\x*200)/0.6) -2 - 0.3*\x});
				\tikzfillbetween[of=A and B]{latentColor, opacity=0.2};
            \end{scope}
    
    \node [trapez, 
        draw=aeColor!80!black, 
        fill=aeColor!20!white,
        pca_opacity2,
        trapezium angle=-50, 
            minimum width=23mm, 
            minimum height=15mm,
            right=4*\sepSpace of latent_trajectory.east,
            label={[label distance=-0.75mm, pca_opacity2, mylabel, rotate=28]above:}
            ] (decoder) {$\decoder$}; 
    

    \node[
        canvas is zy plane at x=3*1,
        inner sep=0pt,
        anchor=south,
        draw=stateColor!80!black,
            ] at ($(decoder.east) + (-19\sepSpace,-0.95)$) (stateReconstr)
        {\includegraphics[height=18mm]{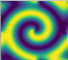}};

    \node[
        canvas is zy plane at x=3*1,
        inner sep=0pt,
        draw=stateColor!80!black,
            ] at ($(stateReconstr.east) + (-33*\sepSpace,-0.175)$) (stateReconstr2)
        {\includegraphics[height=18mm]{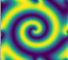}};

    \node [cell, 
        draw=none,
        minimum height=9.5mm,
        minimum width=5.5mm,  
        latent_opacity,
        anchor=east,
        ] at ($(stateReconstr2.east) + (.6,-0.25)$) (dots) {$\mathbf{\dots}$};
    
    \node[
        canvas is zy plane at x=3*1,
        inner sep=0pt,
        draw=stateColor!80!black,
            ] at ($(stateReconstr2.east) + (-18\sepSpace,-0.175)$) (stateReconstr3)
        {\includegraphics[height=18mm]{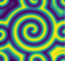}};

    \node[
        canvas is zy plane at x=3*1,
        inner sep=0pt,
        anchor=south,
        draw=stateColor!80!black,
            ] at ($(decoder.north east) + (-19\sepSpace, 0.85)$) (uqReconstr)
        {\includegraphics[height=18mm]{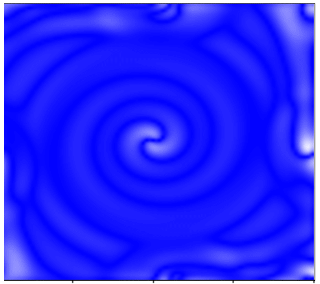}};

    \node[
        canvas is zy plane at x=3*1,
        inner sep=0pt,
        draw=stateColor!80!black,
            ] at ($(uqReconstr.east) + (-33*\sepSpace,-0.175)$) (uqReconstr2)
        {\includegraphics[height=18mm]{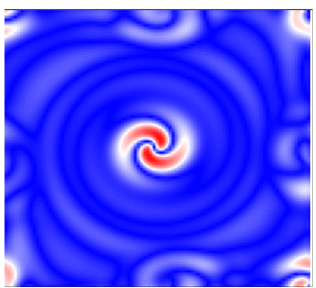}};

    \node [cell, 
        draw=none,
        minimum height=9.5mm,
        minimum width=5.5mm,  
        latent_opacity,
        anchor=east,
        ] at ($(uqReconstr2.east) + (.6,-0.25)$) (dots) {$\mathbf{\dots}$};
    
    \node[
        canvas is zy plane at x=3*1,
        inner sep=0pt,
        draw=stateColor!80!black,
            ] at ($(uqReconstr2.east) + (-18\sepSpace,-0.175)$)  (uqReconstr3)
        {\includegraphics[height=18mm]{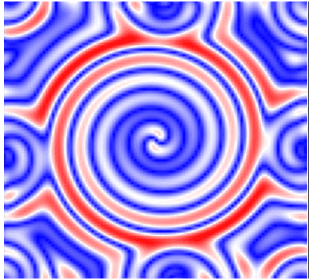}};
        

    \begin{pgfonlayer}{bg1}    
        \node[
            fit=(state) (stateReconstr3) (param), 
            cell, 
            dashed, 
            draw=none, 
            fill=none, 
            inner sep=5*\sepSpace, label={[label distance=1mm, mylabel]above:}] (viciBox) {};
        \node[
            fit=(vindyDist) (latent dist), 
            dashed,
            draw=VariationalColor, 
            fill=VariationalColor!3!white, 
            inner sep=1*\sepSpace, label={[label distance=0mm, mylabel, text=VariationalColor!80!black]below:sampling}] (sampling) {};
        \node[
            fit=(state) (param), 
            dashed,
            draw=stateColor, 
            fill=stateColor!3!white, 
            inner sep=1*\sepSpace, label={[label distance=0mm, mylabel, text=stateColor!80!black]below:inputs}] (sampling) {};
        \node[
            fit=(stateReconstr) (stateReconstr3), 
            dashed,
            minimum height=24mm,
            draw=stateColor, 
            fill=stateColor!3!white, 
            inner sep=1*\sepSpace, label={[label distance=0mm, mylabel, text=stateColor!80!black]below:prediction}] (sampling) {};
        \node[
            fit=(uqReconstr) (uqReconstr3), 
            dashed,
            minimum height=24mm,
            draw=latentColor, 
            fill=latentColor!3!white, 
            inner sep=1*\sepSpace, label={[label distance=0mm, mylabel, text=latentColor!80!black]above:UQ}] (sampling) {};
    \end{pgfonlayer}

    
    \draw [Arrow, shorten >= 2pt, surrogate_opacity, shorten <= 2pt] (latent dist.east) -- +(.75,0) node[pos=0.2, above, rotate=90,] {};
    \draw [Arrow, shorten >= 2pt, surrogate_opacity, shorten <= 2pt] (vindyDist.east) -- +(.75,0) node[pos=0.2, above, rotate=90,] {};
    \draw [Arrow, shorten >= 2pt, surrogate_opacity, shorten <= 2pt] (param.east) -| (ode.north) node[pos=0.2, above, rotate=90,] {};
    
    \draw [Arrow, shorten >= 2pt, surrogate_opacity, shorten <= 2pt] (ode.south) |- ($(ode.south)+(-0.5,-0.75)$) -| (trajectory.north) node[pos=0.45, right, align=left, rotate=0,] {\small time\\ \small integration};
\end{tikzpicture}
